# Federated Learning for Big Data:
# A Survey on Opportunities, Applications, and Future Directions

Thippa Reddy Gadekallu[a,b], Quoc-Viet Pham[c], Thien Huynh-The[d], Hailin Feng[e], Kai Fang[e,1], Sharnil Pandya[f], Madhusanka Liyanage[g], Wei Wang[h], Thanh Thi Nguyen[i]

[a]*The College of Mathematics and Computer Science, Zhejiang A&F University, Hangzhou 311300, China (email: thippareddy@ieee.org)*

[b] *Division of Research and Development, Lovely Professional University, Phagwara, India*

[c]*School of Computer Science and Statistics, Trinity College Dublin, Dublin 2, Ireland (email: viet.pham@tcd.ie)*

[d] *Department of Computer and Communications Engineering, Ho Chi Minh City University of Education and Technology, Vietnam (email: thienht@hcmute.edu.vn)*
[e]*College of Mathematics and Computer Science, Zhejiang A&F University, Hangzhou 311300, China. (Email: hlfeng@zafu.edu.cn, Kaifang@ieee.org)*
[f]*Department of Computer and Information Sciences, Faculty of Engineering and Environment, Northumbria University, Newcastle upon Tyne NE1 8ST, United Kingdom. e-mail: (sharnil.pandya@northumbria.ac.uk)*
[g]*School of Computer Science, University College Dublin, Ireland and Centre for Wireless Communications, University of Oulu, Finland (email: madhusanka@ucd.ie and madhusanka.liyanage@oulu.fi)*
[h]*Guangdong-Hong Kong-Macao Joint Laboratory for Emotion Intelligence and Pervasive Computing, Artificial Intelligence Research Institute, Shenzhen MSU-BIT University, Shenzhen, 518172, Guangdong, China , School of Medical Technology, Beijing Institute of Technology, Beijing 100081(email:e-mail:ehomewang@ieee.org)*
[i]*School of Information Technology, Deakin University, 75 Pigdons Rd, Geelong, VIC, 3216, Australia (email: thanh.nguyen9@monash.edu)*

**Abstract**

In the recent years, generation of data have escalated to extensive dimensions and big data has emerged as a propelling force in the development of various machine learning advances and internet-of-things (IoT) devices. In this regard, the analytical and learning tools that transport data from several sources to a central cloud for its processing, training, and storage enable realization of the potential of big data. Nevertheless, since the data may contain sensitive information like banking account information, government information, and personal information, these traditional techniques often raise serious privacy concerns. To overcome such challenges, Federated Learning (FL) emerges as a sub-field of machine learning that focuses on scenarios where several entities (commonly termed as clients) work together to train a model while maintaining the decentralisation of their data. Although enormous efforts have been channelized for such studies, there still exists a gap in the literature wherein an extensive review of FL in the realm of big data services remains unexplored. The present paper thus emphasizes on the use of FL in handling big data and related services which encompasses comprehensive review of the potential of FL in big data acquisition, storage, big data analytics and further privacy preservation. Subsequently, the potential of FL in big data applications, such as smart city, smart healthcare, smart transportation, smart grid, and social media are also explored. The paper also highlights various projects pertaining to FL-big data and discusses the associated challenges related to such implementations. This acts as a direction of further research encouraging the development of plausible solutions.

*Keywords:* Big Data, Federated Learning, Smart City, Smart Healthcare, Smart Transportation, Security.

## 1. Introduction

Machine and Deep learning methodologies have created a significant impact in a variety of federated learning-enabled big data applications in areas such as preventive healthcare, smart cities, neuroscience, precision agriculture, renewable technologies, etc. The advent of big data technologies has gained new insights into business applications integrated with machine learning methodologies. Big data technologies are capable of extracting and feeding rich information to machine learning methodologies for better decision-making. The extracted rich information can assist machine learning-enabled systems in identifying homogenous and heterogeneous patterns for building sustainable predictive models. The conceptualization of integrating federated machine learning methodologies with big data technologies can enable information processing at multi-granularity levels, managing high-data dimensionality, optimized computing, and model scalability[1, 2]. It is expected that the big data market may reach around 230 billion $ by the year 2025 [1]. Almost all the verticals of industries such as entertainment, retail, finance, media, manufacturing, healthcare, and social media benefit from the big data. Big data has attracted a great deal of attraction from academia and industry in recent years. Big data has a potential to solve many of the challenging problems in several sectors. The big datasets can be analyzed to uncover patterns through data analytics that attracted the researchers. Some of the examples where big data can play a very important role include using the consumer data, radio frequencies, data from social media, and global position systems for improving several sectors in urban life such as energy, trans-

[1]https://www.prnewswire.com/

portation, education, etc. in a smart city, predicting the patients with high risk of cardiac arrests and other health related risks in real time through predictive analytics using the large volumes of electronic health records, etc. Even though big data is used prominently in today's world, there is still a vagueness in the definition of "big data". According to European Commission, big data is large quantity of several types of data generated from different sources like machines, people, or sensors[2]. The data generated may be related to videos, posts in social media, digital images, satellite imagery, and climate information [3]. Big data datasets are of tremendously large sizes with huge dimensions, that cannot be captured, stored, managed, or processed by traditional database technologies. Several characteristics of big data include volume, velocity, veracity, variety [4].

The true potential of big data can be realized only when the value is extracted from massive data by data analytics. Machine learning (ML) plays a very important role in such scenarios due to its ability to understand the patterns and provide insights by learning from the data [5]. ML algorithms can be categorized as supervised ML and unsupervised ML. In supervised ML, the algorithms know both the inputs as well as outputs in prior and the ML algorithms learn how to map inputs to the outputs. Classification and regression are examples of supervised ML. In unsupervised ML, the outputs are not known to the ML algorithms but they discover the patterns within the data on their own. Clustering is an example for unsupervised ML [6]. Traditionally, the data gathered from several big data sources is transferred to a central cloud, where the ML algorithms are trained to understand the patterns from the stored data. This approach of training the big data in a central cloud faces several challenges like increased latency, connectivity, bandwidth, data migration, exposing of sensitive information (privacy preservation). Poor decision making due to latency would reduce the system throughput in real-time analytics. In such cases, edge computing addresses the issue of high latency when transmitting data from a device across a network to a centralised server, reduces operational costs, and increases the performance [7, 8, 9]. Edge computing, on the other hand, faces some challenges, such as privacy and security [10]. In a distributed edge environment, ensuring adequate security can be difficult because sensitive data about people, vehicles, governments, and organisations can be exposed [11]. Edge computing uses a decentralized model to process data. Each node has a high chance of becoming vulnerable. In centralized systems, security methods are highly concentrated, whereas in edge environments, comprehensive and consistent security protocols across all nodes are necessary. In edge computing different nodes have varying levels of security, hardware, firmware updates, and security configurations, making some nodes more secure than others. If data is transmitted between edge devices and central systems without encryption, sensitive information can be stolen. Updating distributed devices in remote locations is a difficult task. These edge devices will not have updated firmware and software, causing them vulnerable to attacks. Insecure APIs used for communication and data exchange among edge devices and central systems have security constraints which can be exploited to gain unauthorized access towards sensitive data.

Threats to validity: For the presented study, we have reviewed studies that deal with federated machine learning and big data technologies, the challenges of integrating big data technologies and machine learning methodologies, big data resource management, sustainable cloud computing for big data, distributed machine learning, security and privacy of federated learning for big data, federated learning for IoT, federated transfer learning framework for wearable devices, communication-efficient federated learning, fuzzy datasets, decentralized federated learning, and FL-enabled autonomous systems[12, 1, 13]. In this survey, we have used multiple combinations of keywords such as 'FL-integrated machine learning', 'big data frameworks', 'distributed machine learning', 'decentralized federated learning', 'federated transfer learning', etc to retrieve peer-reviewed articles on journals, book and conference proceedings, world reports, book chapters, etc. For the review of work, we have targetted six reputed databases, web of Science, Scopus, IEEE Xplore, SpringerLink, Google Scholar, ScienceDirect, ACM Digital Library, etc. The study title and abstract were screened to identify potential matching research articles. The full text of 143 studies were assessed to identify the relevant contents. In total, we have eventually shortlisted 143 articles out of 3012 retrieved documents.

Federated learning (FL) is a recent development in ML, which has evolved from the concept of distributed ML. Distributed ML is a multi-nodal system that has the capability to resolve computational issues faced by big data. The distributed ML benefits the existing big data-based methodologies by providing scalability and efficiency. Distributed ML models will be trained using the data spread across multiple nodes. After the training, the local nodes will transmit the updated weights to the centralized cloud server. Once the centralized cloud server receives the updated weights from the distributed nodes, it aggregates the weights for generating the output of the global ML model. This process will be repeated till the desired results are realized [14]. The distributed ML speeds up the training process of big data exponentially that supports the scalability of big data. Split Neural Networks (SplitNN) is one of the widely known distributed learning algorithms wherein the neural network (NN) training is "split" among two or more hosts. Each segment of the model acts as a self-sufficient NN and is fed into the segment in front of the same.[15].

Federated Learning (FL) is an extension of distributed ML approach. However, FL differs from the distributed ML approach as follows [16]:

- In FL, the transmission of the raw data to the centralized cloud server is not permitted. Whereas, in distributed ML approach, there is no such restriction.
- In a distributed ML, the centralized cloud server belongs to a single organization, whereas, FL enables sharing of the resources across multiple organizations.

[2] https://digital-strategy.ec.europa.eu/en/policies/big-data

- Distributed ML does not take the required measures to handle privacy and security issues. On the other hand, FL uses encryption mechanisms to enhance privacy and security of the raw data.

For the purpose of training the ML algorithms, FL does not send the raw client data to the central model located on the cloud server. Local clients are tasked with handling the global machine learning algorithm itself. The FL architecture contains a centralized cloud server that is responsible for coordinating the training of global ML model across the clients that are traditionally local devices. The clients are trained using the locally generated data by the global ML model. During this process, the weights of the global ML model are shared with the local clients by the centralized cloud sever. After the completion of local training, the updated ML parameters such as skewness, accuracy, loss, etc will be transmitted back to the centralized cloud server for aggregation [17]. This property of FL has a great potential to solve the aforementioned issues in big data. The differences between distributed ML and FL are summarized in Table 1.

Owing to the importance and practicality of FL and big data, there have been a number of surveys on these two topics as summarized in Table 2.

Despite a number of important review articles on FL and big data, a comprehensive survey on the use of FL for big data has not been done yet. To fill this research gap, we aim to provide a concise survey on FL for big data. In particular, we focus on highlighting the opportunities that FL brings to big data services and applications. In summary, contributions and features offered by this study can be summarized as follows.

- **A preliminary to FL and big data**: Firstly, we present the fundamentals of FL and big data and discuss various motivations behind the use of FL for big data services and applications, such as mitigating security and privacy issues, reducing communication cost, handling the issue of data variety, easing the big data analysis, and improving the system scalability. We will detail this part in Section 2.

- **FL for big data services:** FL plays an important role in big data services such as big data acquisition, big data storage, big data analytics, and big data privacy preservation. We review the use of FL for these big data services in detail in Section 3.
- **FL for big data applications:** Next, we review FL-empowered big data in vertical domain applications, including smart city, smart healthcare, smart transportation, smart grid, online recommendation systems, social media, and several miscellaneous applications. This part is detailed in Section 4.

- **A review of practical projects on FL-big data:** Several practical projects on FL-big data are reviewed in Section 5 so as to drive real implementations of the FL-big data solutions.

- **Key challenges and promising directions on FL-big data:** Despite various advantages, the use of FL for big data also faces challenging problems such as bottleneck in communication efficiency, data heterogeneity, statistical heterogeneity, and privacy issues. These challenges are explained in detail in Section 6 along with potential directions of FL-big data.

The study's distinctive contribution is its comprehensive coverage of FL, Big Data emphasizing the distinctions between FL-based systems and classical ML. Additionally, the paper describes the need for such aggregation and explores the various aggregation strategies related to FL and ML. The study's literature clearly outlines FL's contribution to big data services, identifies the difficulties that come with using conventional methods, and emphasizes the advantages of using FL across all verticals. Applications of FL are seen in a variety of fields that touch on every facet of society. In the end, the paper identifies numerous projects that effectively adopt FL and highlights the difficulties that come with doing so.

## 2. Federated Learning and Big Data: An Overview

In Big Data, where huge datasets are generated and analyzed, conventional machine learning suffers scalability, privacy, and efficiency issues. Federated Learning trains models on several devices or servers with local data. In this section, we discover the fundamental concepts of FL and big data. The motivations of their integration is given correspondingly.

### 2.1. Federated Learning

Recognized as a new learning paradigm where statistical models are trained on multiple local devices in a distributed network without sharing their training data, FL provides AI tools to train a model collaboratively by using a federated dataset of secure sources. The rise of smart devices, such as mobile phones, wearables, and autonomous vehicles, has led to local data storage and processing in distributed networks, highlighting the importance of information privacy. The general operation of FL, its classification and related core challenges are emphasized in this study. A general FL architecture with processing flow is presented in Figure. 1. Smartphones, smart vehicles, and mobile computers with local data and privacy-preserving models act as an intrinsic part of the FL architecture. All devices download the global model from FL server, train it locally on their data, and send the new parameters to server. The server collects recent modifications to improve the global model, which is transmitted to all clients. Each round improves the global model by not transferring raw data between devices and the server to protect data privacy. Time complexity is a critical factor in the performance of AI models, particularly in federated learning environments. The time complexity of AI models affects the efficiency and feasibility of deploying these models in real-world applications [28].

To understand and differentiate the diverse approaches and configurations of FL frameworks, we categorize them based on two key aspects: training data property and computing paradigm.

Table 1: Comparative analysis of big data services in conventional distributed ML vs FL.

| Big Data Services | Conventional Distributed ML | Federated Learning |
|---|---|---|
| Big Data Acquisition | Yields better performance while handling balanced, and homogeneous data. | Yields better performance while handling unbalanced, and heterogeneous data as well. |
| Big Data Storage | 1. Employs a single server or a cluster in a single organization. 2. Due to activation of multiple devices, relatively low tolerance and robustness is observed. | 1. Utilizes distributed computing resources in multiple clusters, regions, organization. 2. Due to activation of only fraction of devices, high fault tolerance and robustness is observed. |
| Big Data Analytics | Recommendation and prediction results are promising at the node level alone. | Recommendation and prediction results are promising at the node as well as centralized level. |
| Big Data Privacy Preservation | Raw data is communicated directly which may result in data leakage. | Raw data is not allowed to communicated directly which ensures privacy preservation. |

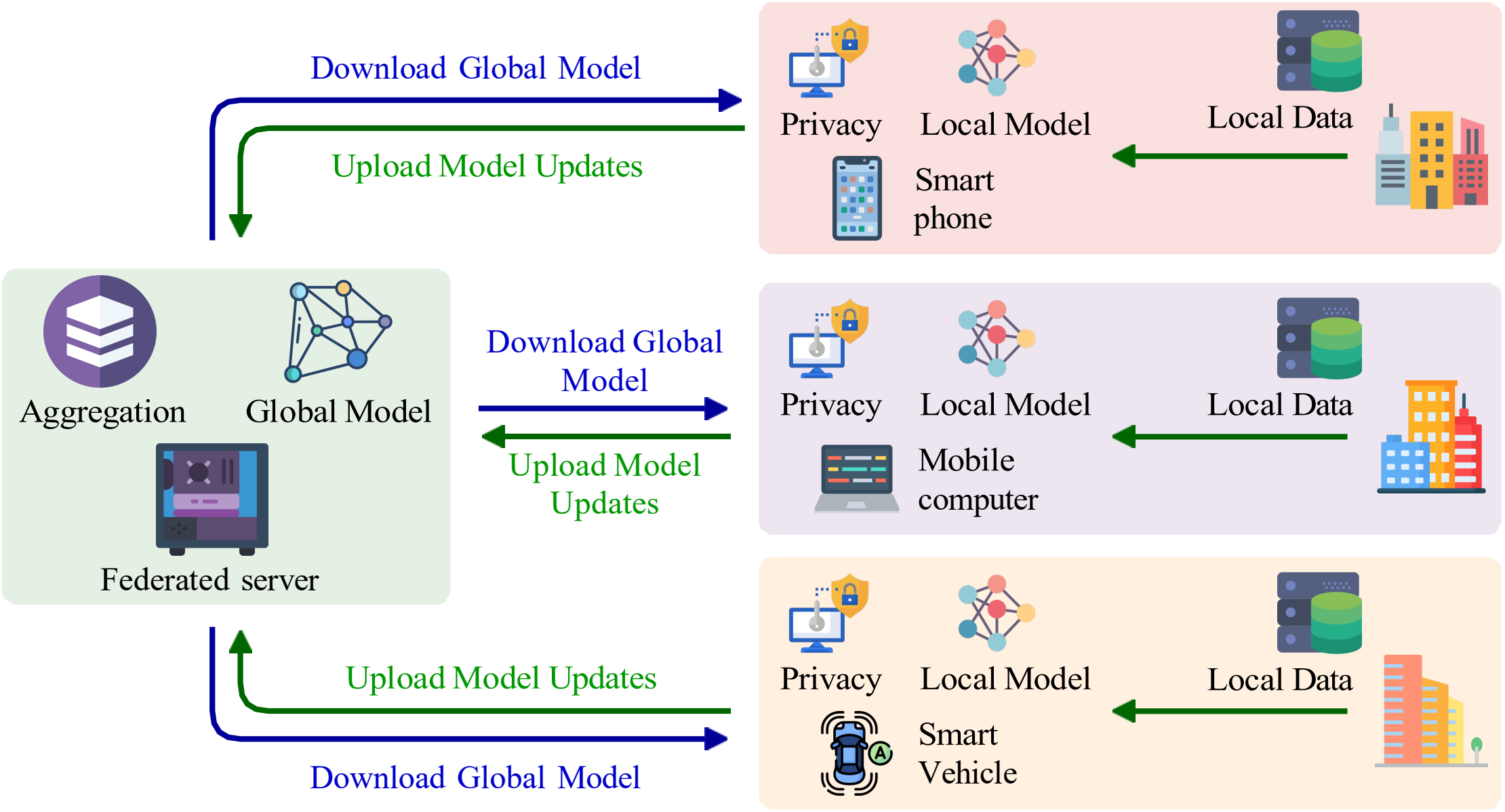

Figure 1: General architecture of FL with processing flow.

Training data property refers to how data is distributed across devices, which can be horizontal (similar features, different samples), vertical (same samples, different features), or a combination through federated transfer learning. Computing paradigm, on the other hand, focuses on the communication and computation architecture used for training models, which encompasses centralized, decentralized, or hybrid approaches. With these categorization strategies, we can analyze the strengths and weaknesses of different frameworks depending on specific applications and data distribution scenarios. By examining these two aspects, we can effectively select the most suitable FL framework for a particular task, ensuring efficient and privacy-preserving model training.

*Training data property*: The characteristics and distribution of training data across different clients in a federated learning system, categorized based on the alignment and overlap of feature spaces and sample IDs into three sub-classes, including horizontal FL (HFL), vertical FL (VFL), and federated transfer learning (FTL), based on the properties of training data [13, 29].

- Horizontal FL: The local devices in HFL frameworks use datasets that have the same feature space but separate sample spaces to train statistical models. Due to this property of input training data, all devices should utilize a same AI model (e.g., support vector machine and deep neural networks). For an example of HFL applications, the hospitals collect the health data of different patients with the electronic medical records having the same at-tributes.

Table 2: Summary of related review papers on federated learning and big data.

| Ref | Contributions | Limitations |
|---|---|---|
| [7] | This work presents two important aspects of big data: deployment requirements and network environments | The use of AI and FL for big data services and applications is not presented. |
| [18] | A preliminary to FL is presented, followed by a thorough review of FL-enabled solutions for privacy, resource allocation, and applications to mobile edge networks. | This survey only focuses on the use of FL at mobile edge networks. |
| [19] | A comprehensive survey of FL for IoT services and IoT applications is presented. | This survey only presents the use of FL for IoT services and applications. |
| [20] | A survey on fusion of FL and IIoT (e.g., background, data management, and IIoT applications) is presented. | This survey does not focus on the use of FL for big data. |
| [21, 22] | The sources of privacy/security/threat issues, types of attacks, unique features, and potential defensive solutions for FL. | These works are only limited to security/privacy/threat aspects of FL. |
| [23, 24, 25] | The use of FL for particular applications, e.g., healthcare, vehicular IoT, and open bank are presented. | These studies do not focus on the role of FL for big data services and applications. |
| [26] | A concise review on background, key techniques, and important applications of big data | This survey was done before the invention of FL, and thus lacks the role of FL for big data services and applications. |
| [27] | A comprehensive survey of big data for intelligent transportation systems, including big data characteristics, applications, and platforms. | This survey only focuses on machine learning for big data modeling and analytics, while the use of FL is ignored. |
| Proposed Work | Comprehensive survey on the use of FL for various big data services and applications, a detailed study of the fundamentals and motivations of integrating FL into big data, recent developments, provided a comparative study of FL-enabled big data applications, existing challenges, and their benefits in various application areas, highlighted several key challenges of integrating FL into Big data and discussed possible solutions and directions for future research work. | |

- Vertical FL: In contrast to HFL, VFL is presented for the cases where the local datasets have different features but the same samples. To collect overlapping data samples in different local devices, an entity alignment mechanism is utilized. For an example of VFL scenarios, different hospitals collect the health data of the same patient but differ in attributes, such as diabetic retinopathy in hospital A, cardiovascular disease in hospital B, and diabetic nephropathy in hospital C.

- Federated transfer learning (FTL): Federated transfer learning is a novel strategy that combines the combines the concept of transfer learning and federated learning. Federated learning is a machine learning environment in which a large number of clients (such as mobile devices or entire organisations) work together to jointly train a model while maintaining decentralised training data, all under the supervision of a central server. Conversely, transfer learning is a technique in which a model created for one job is used to another task as the foundation for a new model [30]. Basically, FTL permits collaborative model training across several activities, organizations or domains ensuring protection of privacy and confidentiality. This is extremely beneficial in industries such as finance and healthcare. As an example, in the healthcare sector, the successful FL model employed by one hospital yielding satisfactory results in the diagnosis of a disease can be further used by multiple hospitals for diversified disease diagnoses. In this way, FTL can enable collaboration across various healthcare organizations beyond the limit of geographical boundaries.

*Computing paradigm*: The architectural approach in the FL system defines how participants coordinate and communicate during the learning process, which can be centralized, with a single coordinating server, or decentralized, with direct communication and coordination among participants [31].

- In a centralized FL system, all devices transmit the trained parameters of local models to the FL server for updating the global model with a parameter aggregation algorithm. Later, the sever delivers the computed global model to all clients in a network for the next training iteration [32].

- In a decentralized FL system, all devices are connected over peer-to-peer communications. A device can transmit the model parameters, locally trained on its own dataset, to neighbors and receive their model updates to aggregate parameters.

After categorizing FL based on training data property and computing paradigm, understanding the mechanisms that enable these categories to function is essential to build and deploy an effective FL framework. As an important component of FL, regardless of whether the FL system is horizontal, vertical, or transfer learning-based, and whether it follows a centralized or decentralized computing paradigm, aggregation performs the model update process for a global model robustly from multiple locally trained models. In any FL frameworks, the aggregation algorithms play a vital role to attain the efficiency, security, and accuracy of the FL process. In the following section, we briefly introduce various aggregation algorithms specifically designed for FL. Based on the concept of collecting the updated weights after training the ML algorithm with the data from the local devices to the centralized cloud server to envision the global ML model [33], some of the advanced aggregation mechanisms in FL are discussed below:

- **Federated Averaging (FedAvg):** FedAvg begins with the random initialization of an ML model on the centralized cloud server. In FedAvg, the global ML model averages the weights obtained from the local devices, which are averaged by the centralized cloud server. The outcome of ML model is sent to local devices to resume the training [32].

- **Federated Personalization (FedPer):** In FedPer, the calculation of the new weights in the aggregated models is similar to that of FedAvg method. The FedPer method differs from the FedAvg in the following ways: (i) how different parts of the ML model are considered during the aggregation process, (ii) only base layers of the clients are communicated to the centralized cloud server. Other layers are retained by the clients. Hence, in FedPer, the decision-making process is carried out by the upper layers of the concerning clients [33].

- **Federated Match Averaging (FedMA):**FedMA matches and averages hidden elements (i.e., convolution layers channels, LSTM hidden states, and neurons for fully connected layers) with similar feature extraction signatures to build the shared global model layer by layer. In FedMA, each layer has an aggregation procedure that enables a set of nodes in that layer to be utilised to address a sub problem. As a result, FedMA can carry out the aggregation seamlessly. [34].

- **Federated Stochastic Gradient Descent (FedSGD):** The FedSGD is a widely known aggregation method that is specifically designed to handle the issue of heterogeneity by minimizing the training loss function [35].

- **Federated Distance (FedDist):** The FedDist is a fusion of FedAvg and FedMA methods. The FedDist method emphasizes more on calculating the distance between the neurons using euclidean distance method. Furthermore, the FedDist method is more suitable for sparse data, and hence assist in building more generalized ML model [33].

The deployment of a single, generalized FL model can be challenged by scenarios with non-identical and independent distributions (non-IID) of data at each client, that means, an universal approach might not be optimal. Federated Neural Architecture Search (FedNAS) [36], a promising automates the FL process by collaboratively searching for an optimized neural network architecture across clients. By combining hyperparameter tuning with FedNAS, we can leverage the strengths of both data property and computing paradigm categories. Even within a broader FL architecture defined by these categories (e.g., HFL with centralized server), personalization techniques can address non-IID data distributions, leading to a more robust and adaptable FL system.

### 2.2. Big Data

Nowadays, a massive data volume are generated daily with unprecedentedly increasing rate from heterogeneous sources. This is due to the emergence of the internet of things, the proliferation of the cloud computing, and the spread of smart devices. Big data refers to massive datasets with heterogeneous formats: structured, unstructured, and semi-structured data, which has three key characteristics: large volume with the incessant generation from millions of devices and applications, the high velocity with the real-time data acquisition and processing, and high variety with various source and multiple formats.

**Taxonomy of big data:** Here we present a taxonomy of big data in various perspectives, including data domain, storage infrastructure, compute infrastructure, and AI-based data processing [37]. This taxonomy is illustrated in Figure. 2.

**Data domain**: Various domains, in which the use of big data has increased tremendously in the last decade, can be determined based on the time span of processing and the degree of data structure. For the time span of data processing analysis, data domain can be classified into three groups: batch (including bioinformatics, geosciences, forensics, and other large-scale sciences), near real-time (including retail and sensory data), and real-time (including finance, social networking, and network security) [38]. We can also classify various domains into three degrees of structure: structured data - containing relational databases and spreadsheets (including retail, finance, bioinformatics, and geosciences), unstructured data - containing raw signals from nature or data without pre-defined structural model (including voice, image, video, and sensory data), and semi-structured data - as the hybrid form of structured and unstructured data (including web logs, email, and documents). Notably, relying on specific applications, the visual media domain with high-dimensional unstructured data can spread from batch to real-time of time span [39].

**Storage infrastructure**: Numerous big data storage solutions have been introduced to engineers and developers when they build services and applications running with large datasets. In Figure. 2, we can find the taxonomy of different types of databases being used for big data storage. The storage platforms in the relational structured query language (SQL) group are with some features: speed over scale, large capacity for vertical scaling, good consistency over availability, and easy deployment. Inspired by relational SQL, newSQL offers higher performance and scalability, thus enables high throughput online transaction processing requests while maintaining high-level language query capability. In the nonSQL group, we have several database techniques for specific purposes, such as document oriented, graph oriented, key-value storage, and big table inspired. Although numerous database platforms have been introduced for diversified services and applications with different data types, they should comply with some standards, for example ACID (atomicity-consistency-isolation-durability) and BASE (basically available-soft state-eventually consistent).

**Compute infrastructure**: Based on the time span metric, computing paradigms for big data can be classified into two primary categories: batch mode and real-time/near real-time mode. For the batch mode with high processing latency, MapReduce and Bulk synchronous parallel are recommended for large parallel and distributed systems. Whereas low-latency computing frameworks for streaming demand immediate response to every incoming event. Some popular frameworks in this group are Infosphere, Apache Storm, and Apache Spark.

**AI-based data processing**: Artificial intelligence (AI) techniques and machine learning (ML) algorithms allow analyzing and processing data in automatic and scalable ways, in which the high-level information and meaningful context can be derived from raw data. ML allows computers to learn complicated patterns automatically and make decisions of new incoming data events [40]. Depending on data utilization of model learning, these algorithms can be categorized into four classes: supervised learning, unsupervised learning, semi-supervised learning, and reinforcement learning. Recently, deep learning has emerged with recurrent neural network, convolutional neural network, and self-organizing map, which has achieved remarkable success in various domains due to the great capability of dealing with large noisy-messy-confusing datasets [41].

### *2.3. Motivations of the Integration*

With the unprecedented growth of IoT services and applications, traditional AI/ML techniques based on centralized computing platforms have exposed several issues: security, privacy, variety, communication, analysis, and scalability. With the ability to train statistical models on local devices and then transfer the learned parameters to a central server for aggregation, FL is recognized as a promising solution to address the challenges of big data, such as privacy concerns and data type diversity [42]. The main motivations behind such kind of integration are summarized as follows:

- **Security and data privacy**: To extract meaningful information and capture patterns of the data generated by local devices, traditional AI and ML algorithms are usually adopted in the centralized server to train statistical learning models. To this end, the data should be transferred from devices to the server for training a model, which in turn exposes some security and privacy risks over cyberattacks. Instead of uploading data to the server, FL allows devices to train models with their datasets locally and transfer model parameters to the server for aggregation.

- **Communication cost**: A massive volume of raw sensory data and low-level informative data of clients should be transferred to the centralized server for processing, which consequently suffers a considerable communication cost, especially with high-dimensional unstructured data like images and videos. With FL, the communication cost will be reduced significantly owning to transferring trained statistical models to the FL server instead of the entire data.

- **Data variety**: In several integrated services and applications, the multimodal sensory data acquired from various sources has different types and structures (structured data vs unstructured data and time series data vs high-dimensional data), which in turn makes difficult to manage in the big data server deploying a specific storage platform. Besides, each data type characterized by various attributes will demand a specific computing infrastructure to achieve best performance of target services and applications. With FL, each devices and clients may take into consideration a specific data type locally, thus reducing the complexity of local data management at the FL server.

- **Analysis**: In many centralized computing systems, learning classification and regression models with the multimodality data can be accomplished using advanced methods (e.g., data-based, feature-based, and decision-based fusing mechanism), but its processing framework will be much more complicated. Whereas FL can offer a universal DL model to different data modalities of a same learning task.

- **Scalability**: In conventional centralized systems, the scalability of storage and computing infrastructures at the server is problematic because of the complex characteristics of big data with 5Vs (i.e., veracity, value, volume, variety, and velocity). In the FL platforms, the server only performs an aggregation function with multiple locally trained models to modernize the global model, hence system scalability is capable of performing at local devices. Intuitively, the scalability of FL is more flexible and convenient than that of centralized learning in the server.

### *2.4. Contributions of the Presented Study*

The contributions of the presented study are as follows:

- In this study, we have discussed a comprehensive survey on the use of FL for various big data services and applications in areas such as preventive healthcare, smart cities, neuroscience, precision agriculture, renewable technologies, etc.

- In the presented review, we have provided a detailed study of the fundamentals and motivations of integrating FL into big data, recent developments, and the difference between conventional distributed machine learning and federated learning methods.

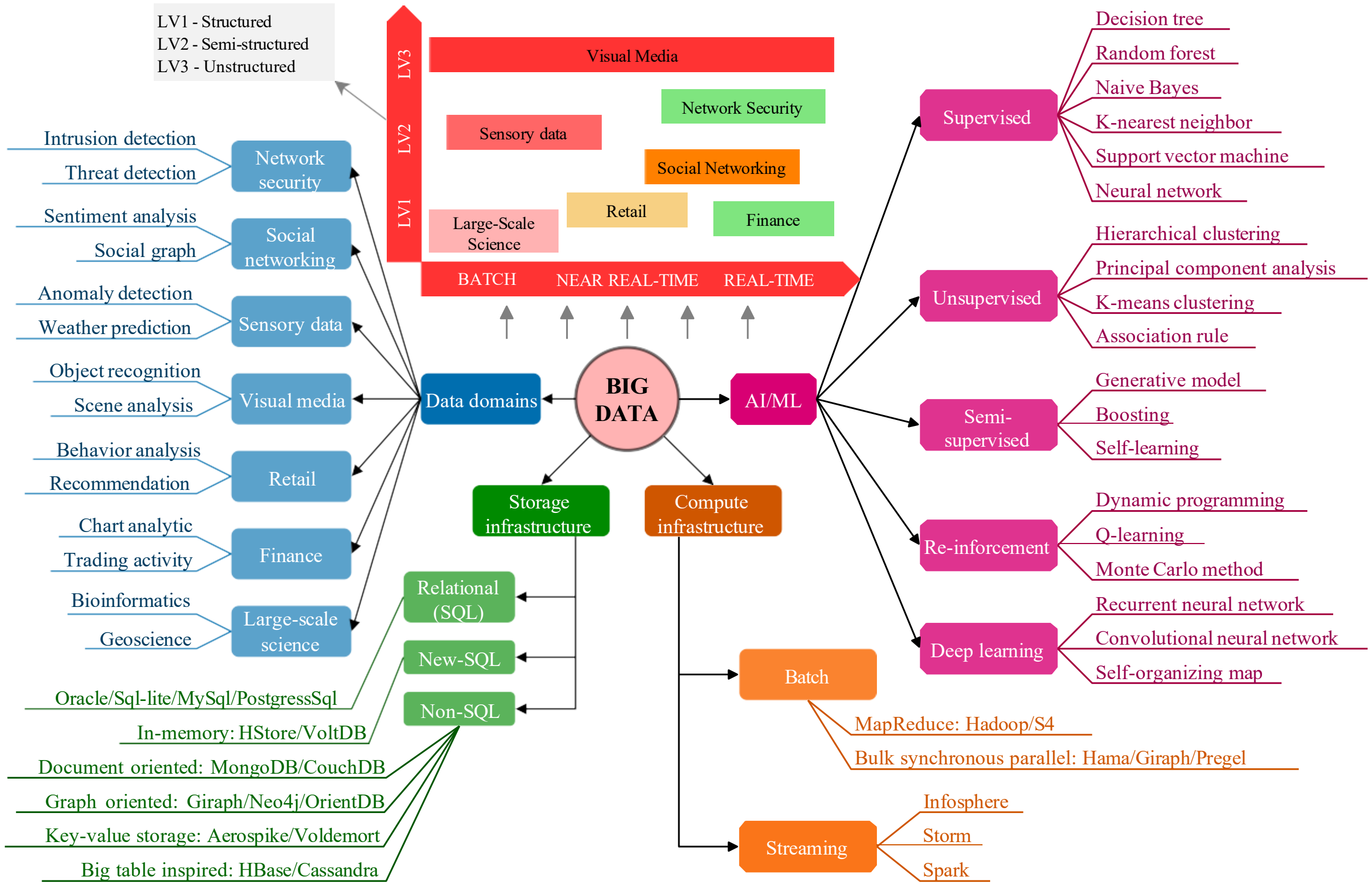

Figure 2: Taxonomy of big data technology.

- In the conducted study, we have provided a comparative study of FL-enabled big data applications, existing challenges, and their benefits in areas such as healthcare, smart cities, smart transportation, and smart grid.
- We have highlighted several key challenges of integrating FL into Big data and discussed possible solutions and directions for future research work.

## 3. Federated Learning for Big Data Services

FL can play a very crucial role in big data services such as acquisition of big data, storage of big data, big data analytics, and privacy preservation of big data. In this section, besides giving a brief comparison in terms of specification between traditional distributed learning and FL [43] in Table 3, we discuss the role of FL for big data services along with recent state-of-the-art. Figure. 3 illustrates FL for big data services.

### *3.1. Federated Learning for Big Data Acquisition*

The exponential growth of AI, ML, and big data analytics has a major impact on a wide range of applications, including health care system, smart grid, smart cities, intelligent transport systems, and so on. Due to the rapid growth of the applications, large volumes of data are being generated at an exponential rate, encouraging the applications to generate the data at a rapid pace. Moreover, some companies and organisations employ advanced artificial intelligence (AI) approaches and big data analytics to quickly identify consumer behaviour in order to increase productivity, sales, and profits. Data sharing is a key problem in AL and ML to fulfil these requirements. Acquiring public data and performing data analytics, is not a perfect solution since public data has constraints such as poor data quality, inconsistency, and unstructured. Most of the companies prefer not to share their data for a variety of reasons and concerns, such data privacy, data sharing rules, and a lack of information. Some times the companies may occasionally share information with third parties in order to facilitate growth and mutual data cooperation agreements. The primary goal of FL is to train ML algorithms across other distributed devices using its own private and local data. The locally trained models are first transmitted to the server, and then all of the shared models are combined with other updates to build a better global model, which is then shared with all devices. The main advantage of FL is that the training data is held on the device, with individual updates sent to the cloud.

#### *3.1.1. Federated Learning for Secure Big Data Collection*

The most important aspect of FL is that it helps in training shared statistical data on the basis of decentralized devices or servers in a local data set. Although the data scientists and researchers use similar model to train the data, the use of FL eliminates the need to upload private data to the cloud or the exchanging of data with other members in the scientific community. In case of traditional machine learning approaches, where the data resides on a single server, FL reduces the challenges associated with achievement of data security and privacy by maintaining the storage of local data. FL has been extremely successful in handling the various challenges relevant to user privacy protection thereby achieving data security. In FL, the data provisioned at the end-user systems is decoupled and the machine learning models are aggregated ensuring involvement of network parameters in the implementation of deep learning at the centralized server. The primary objective of FL is to understand and learn the global model without compromising with the data privacy. In the FL environment, data manipulation and analysis take place close to where the data is generated, which improves data accuracy. Local data processing reduces data corruption during transmission, thereby maintaining data quality.

Table 3: Specification Comparison between Conventional Distributed ML vs FL.

| Settings | Conventional Distributed ML | Federated Learning |
|---|---|---|
| Generalization | A learning model is trained on a large but "flat" dataset, in which the clients are defined as computing nodes in a single cluster or datacenter. | The clients can be different organizations (e.g., hospital or bank) in cross-silo cases or numerous mobile/IoT devices in cross-device cases. |
| Data distribution | Data should be centrally stored and shuffled across clients over a balancing mechanism. | Data should be generated locally and remains decentralization. Each client stores its own data for use and cannot read the data of other ones. Data can be independently and identically distributed (IID) and non-IID. |
| Orchestration | Centrally orchestrated. | A central orchestration server take responsibility for training the global model, but cannot manipulating raw data. |
| Distribution scale | Typically up to 1000 clients. | Massively up to $10^{10}$ clients (as devices). |
| Data partition | Data can be partitioned/re-oraganized across clients arbitrarily. | Data can be example-partitioned (horizontal FL) and feature-partitioned (vertical FL). |
| Client reliability | Few failures. | Excellent cooperation amongst clients taking part computational iterations (due to certain battery, connectivity, and inactivity requirement concerns)). |
| Client statefulness | High statefulness (each client should participate in each computation round and carry state from round to round). | Low statefulness (each client likely participate only once in a task). |
| Client addressability | Each client should have an identity or name for accessibility by the system. | Clients cannot be indexed directly (no client identification). |

It has unique abilities to achieve data privacy while training the data in a dataset. In case of "anonymized" dataset stored in a server, there are possibilities of client privacy getting compromised if the dataset is linked to other datasets in the framework. On the contrary, in case of FL, the information transmitted has minimum updates which contributes in improvement of accuracy in a machine learning model. One of the studies conducted by [44] used a Verifiable FL (VFL) to enable preservation of privacy of big data in an Industrial IoT (IIoT) setting. Although the traditional deep learning techniques have been quite successful in classification and prediction-related problems, they have potential vulnerabilities related to security and privacy issues. In big data-enabled IIoT systems, the deep learning methodologies are applied on the data collected from the local devices. In conventional FL systems, the shared gradient data contains sensitive information which may result in the leakage of the data at the centralized cloud server. In a VFL setup, instead of gradient data aggregation, the Lagrange interpolation methodology is applied to validate the accuracy of the aggregated gradient data. The experiment was carried out on the MNIST dataset, which contains 70,000 images with 28*28 pixels. The experimental results show that the proposed model achieves 96% accuracy with optimal gradient encryption overhead for each participant. The decentralized approach reduces the need for data to travel to central servers, which minimizes exposure to potential breaches and enhances data security. Another example of FL implementation in big data security is found in [45]. It is a known fact that the credit card transaction datasets are generally skewed and the number of data relevant to fraudulant activities are quite less. As a result, it becomes quite challenging to recognise and evaluate fraud tendencies in order to eventually identify actual frauds. The study proposed a FL for Fraud Detection (FFD) framework which enabled the banks to learn the fraud detection model by training the data distributed in their localised database. The FDS is constructed by the aggregation of local computational updates of the fraud detection model. The banks thus can achieve data privacy without having to share the data or sensitive information of the card holders.

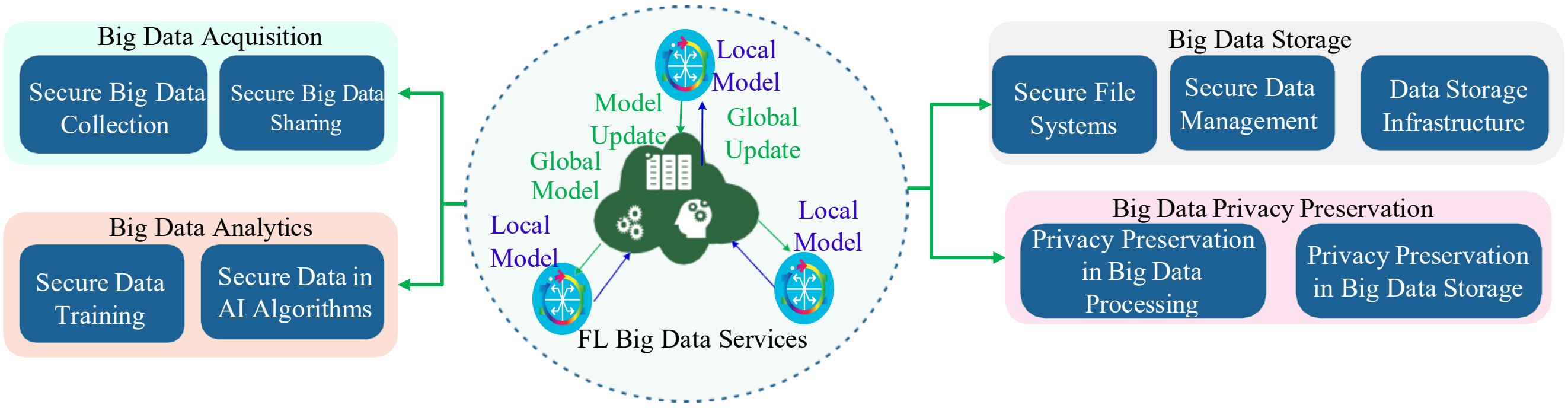

Figure 3: FL for big data services.

The experiment was conducted on a skewed dataset, and the results show that the FL-based FFD achieves 95.5 % accuracy. By processing the data locally and only transmitting model updates, these FL methods reduce bandwidth requirements and improve the efficiency of data acquisition processes. This is particularly valuable in environments with limited connectivity or where large datasets are involved.

*3.1.2. Federated Learning for Secure Big Data Sharing*

Data sharing in big data applications helps in transferring data in a shared network in order to help end users access and use various applications. FL eliminates the process of sharing raw data and takes an alternative approach of sharing the learned results for preserving privacy and ensuring low latency of applications. The study in [46] discusses a integrated data sharing model implemented for industrial IoT systems. To achieve fast and secure data exchange between owners and requesters, the model is configured with a decentralized sharing mechanism for risk reduction of data leakage and access management of parties. The experiment was carried out on two different datasets, one containing 15732 files and the other containing 20000 files. The proposed model achieves 91% accuracy and shows promising results in terms of efficiency and security when compared to other existing models. Furthermore, the approach enhances the data acquisition process in industrial IoT by enabling the collection of knowledge from distributed sources without sacrificing its accuracy and privacy. The study in [47] implemented a framework for a scenario based on industrial IoT, in which FL is operated with centralized servers to to improve data security. In this work, various factories collaborate with the tensor mining to share their data which are encrypted using homomorphic encryption technique. The research was conducted on two industrial datasets in Beijing and Shanghai. The size of the Beijing tensor is 100*600*3, while the size of the Shanghai tensor is 36*150*6. The preliminary results were carried out on three metrics mining quality, eavesdropper security, and stalker security, and provide promising results when compared to other existing models. In terms of data acquisition, the proposed approach securely aggregates processed data from various industrial sites without direct access to raw data. This method improves data security and privacy while also improving data acquisition quality by using knowledge from multiple, geographically distributed sources. This model demonstrates how FL can help obtain valuable analytical insights while adhering to strict security and privacy standards, thereby improving the data acquisition process in important industrial environments. In another interesting work, the authors in [48] presented an asynchronous federated data sharing architecture for internet of vehicles. In this case, the vehicle is an FL client shares the data with the aggregation server located at macro base station (MBS). The vehicles with various service requests pertinent to prediction of traffic or path detection post their requests to the MBS. The MBS implements computing tasks to resolve these requests using an actor-critic reinforcement learning framework. Thus, the system segregates good and bad nodes, enables secured and intelligent data sharing, decision making, thereby ensuring cost optimization. The model's design enables the dynamic collection of data from a distributed vehicle nodes, each of which functions as a separate data generator. This decentralized approach to data acquisition is vital in environments such as the IoT, where data privacy and low latency are essential. By processing data on the vehicle and only sharing model updates with the MBS, the system effectively aggregates knowledge without disclosing raw data. This method not only protects the confidentiality and integrity of data, but it also uses a variety of data sources to improve the overall model's robustness. The proposed asynchronous FL model was tested on the MNIST dataset using a Convolutional Neural Network for local training. According to the experimental results, the proposed model achieves 96% accuracy, optimal convergence rate, and optimal loss with varying numbers of data sources.

*3.2. Federated Learning for Big Data Storage*

In this subsection, The role of FL in storing big data is discussed.

*3.2.1. Federated Learning for Secure File Systems*

In traditional FL, due to various factors such as network latency, data heterogeneity, and resource constraints, if the central server becomes unavailable, the entire training process will be disrupted. The server should also have high-bandwidth and reliable communication links with the agents to transfer big data. Also, all the agents should trust the server. To overcome the aforementioned issues, a decentralized approach, where the model relies on the own resources of the node, can be used as an alternative to perform the training. Pappas et al. [49] proposed a decentralized FL framework, called, interplanetary learning system, that is inspired by interplanetary file system. The proposed framework allows mobile agents to collaborate in the training of the model without relying on any central server. This model was further evaluated on the MNIST dataset, which contained 60000 images, yielding 97% accuracy rate.

#### *3.2.2. Federated Learning for Secure Data Management*

Due to the latest industrial revolution, the user equipments are distributed and widespread. Conventional network infrastructures are unsuitable for Industrial IoT applications due to communication delays and transmission media costs. Several Industrial IoT applications like driverless cars, intelligent robots, smart medical systems, and smart grids are based on wireless networks that generate big data [50]. Many researchers are working on efficient storage, usage, and management of big data generated from Industrial IoT [51]. The big data generated from Industrial IoT often contains private and sensitive information that should not be exposed as it is vulnerable to attacks from malware, heterogeneous equipments or heterogeneous networks [52]. FL can address the aforementioned challenges in the management of big data generated from Industrial IoT devices as it preserves privacy of the users' data by not allowing the global ML model to have access to the sensitive information [53]. FL can also maintain the heterogeneity of the data, hence reducing the deviation of training in ML model. To reduce the training rate of ML model on big data from Industrial IoT and to reduce the model aggregation's communication cost, Zhang et al. [54] proposed a framework based on FL assisted by deep reinforcement learning. In this framework, FL is configured in a centralized system with a novel central server-side agent that takes responsibility of data alignment and client selection for every learning and communication round. The proposed framework was validated on the MNIST dataset and the Fashion MNIST dataset. The authors used deep neural network mode to train the data and achieved a 97% accuracy rate. In a similar work, Kim Sungwoork [55] proposed an incentive mechanism for attracting several owners of data for joining in the process of FL to preserve privacy and efficient management of data. The suggested method uses two ideas: differential privacy to protect privacy and mechanism design to create incentives to meet the goals. The FL framework includes 50 IoT devices, 10 learning operators, and 100 simulation runs. The analytical findings demonstrate that the FL platform performs better than other current models thanks to its high throughput. Real-time data handling is crucial for patient monitoring from a distance. The care professionals must receive the processed data from the patient's big data in order to make judgments quickly. Real-time data processing may give rise to several challenges, including but not limited to patient mobility and computational complexity. When combined with technologies like cloud and fog computing, FL can work well to address these problems. ML-based methods [56].

For handling the challenges and open issues in big data streams generated through IoT devices in smart cities in terms of processing of big data and networking, Imteaj et al. [57] proposed a distributed sensing scheme that paves the way for FL for distributed decision making. In this scheme, a device can be identified by a unique token that is used to activate the user's device. The distributed user's devices transmit data to the centralized cloud server and store it in an appropriate format. The proposed approach ensures remote data collection using the available end-user devices. It also helps in reducing the overall cost as the installation of new sensors is not required.

#### *3.2.3. Federated Learning for Data Storage Infrastructure*

The standard machine learning techniques require training of the data in one machine or a data center. FL enables devices to learn from a shared prediction model ensuring that all the training data is stored in one device. It decouples the ability to implement machine learning based on the need to store data in the cloud. As an example, when a device downloads the current model, it improvises the same by learning from the data in the device and then summarizes the changes in the data as specific update to the model, which alone is sent to the cloud. In this process, encrypted communication technique is used wherein it gets immediately averaged with other updates that initiates improvement of the shared model. Thus all the training data remains on the local device and no updates get stored in the cloud. Thus it eliminates the need to store data in the cloud and also incorporates the use of "Secure Aggregation Protocol" using cryptographic techniques such that a coordinating sever is able to decrypt only the average update when hundreds and thousands of users have participated. It does not allow any individual update to be inspected unless averaging is done [16, 58].

### *3.3. Federated Learning for Big Data Analytics*

The increasing growth of IoT applications generates vast sensor data. Understanding raw data patterns may reveal vital information in created data. Big data analytics can benefit from ML and related technologies. To execute ML services, big data analytics require participant sources to provide their personal and sensitive data to a central server, which may raise privacy concerns for mobile users. FL can efficiently reserve privacy while analyzing huge data. [59]. In this subsection, the application of FL for secure data training and securing the data in AI algorithms is discussed.

#### *3.3.1. Federated Learning for Secure Data Training*

The current approach of performing big data analytics in the cloud has concerns related to network cost and data privacy. FL can address these challenges by training the data in local devices and updating the global model with the local parameters without sharing the raw data. The recent state of the art FL for secured training of big data is presented below.

Xu et al. [60] have proposed a novel framework based on FL, named FL-Pruning, Quantization and Selective Updating (PQSU) to address the network connectivity issues and limited computational resources for performing big data analytics in IoT devices that slows down the training process. The proposed method consists of 3 phases, namely, structured pruning, quantization of weights and updating selective, that col-laborate to reduce the storage, communication and computation costs, thereby fastening up of the training process of big data from IoT devices. The experiment was carried out on the MNIST dataset, which contains 60000 training and 10000 testing grayscale images, and the CIFAR10 dataset, which contains 50000 training and 10000 testing colored images. According to the experimental results, the proposed model achieves 98.9 percent accuracy. In a similar work, to address the problem of leaking of sensitive information of the participants to the untrusted servers in big data analytics, Zhang et al. [59] proposed PEFL, a privacy-enhanced FL approach for protecting gradients in a server which is untrusted. In the proposed scheme, a Paillier homomorphic cryptosystem has been applied to provide protection to preserve the privacy of the users and mitigate information leakage issues, which achieves the secure aggregation at the centralized cloud server. For reducing the cryptosystem's costs of communication of the the authors have used a distributed selective stochastic gradient descent approach for achieving distributed encryption during the training phase. The encrypted gradients enable secured aggregation in server. Hence, the untrustred server can learn only the statistics of the participants that are aggregated, while preserving the privacy of the individuals. In another work, Doku et al. [61] proposed a FL and blockchain based model to retrieve relevant data from big data by using proof of common interest mechanism to segregate relevant data from big data. This data is then trained on a federated model on the local devices. A shared model is then generated by aggregating the model along with other models in the network, which is then stored in the blockchain. The network members can download the aggregated model that can be used to provide edge intelligence to the users.

In a smart grid environment, massive data is generated from the households and industries. Smart metering systems in smart grid have to perform data analytics quite frequently (may be once in every 15 minutes) to understand the patterns in consumption of electricity by the clients, that can enable coping up with the usage of electricity during peak times, seamless integration of renewable sources of energy through coordinate consumption of electricity. Even the retailers can be benefited by big data analytics performed on the data generated by smart meters to have more transparency of consumer's behavior in consuming electricity, through which personalized services can be provided to them [62]. In the retail market, several retailers own the data from smart meters. They may not share the data to the central cloud for big data analytics due to privacy concerns. Hence, the identification of the characteristics of the consumers is difficult as full dataset can be accessed by the global ML models. To address this issue, Wang et al. [63] proposed a centralized FL based approach to train the learning model on the local data residing at diversified clients without sharing the raw data to the global Ml model, thereby preserving the privacy of the data. This work adopted a horizontal FL framework; wherein principal component analysis is applied to extract the significant features of the same space from different sample space for learning local models at clients. Artificial neural networks based ML algorithms are trained on the consumer data using FL approach for big data analytics. The experiment was carried out on the CER dataset, where 80% of the data was used to train the identification model and the remaining 20% was used to verify the efficiency. The results show that the proposed model achieves 94% accuracy. In a similar work, Zhai et al. [64] proposed a centralized FL framework, constrained by delay deadline that overcomes long delays due to training of big data generated in smart grids. Remarkably, two AI-aided client selection algorithms were developed based on the concepts of client maximum greedy and utility-positive guarantee to optimize computing utility in different applications. The experiment was carried out on MNIST, which contains 60,000 training images and 10,000 testing images. In terms of computing utility, the simulation experiments showed that the proposed model outperforms FedCS and cli-max greedy.

#### 3.3.2. Federated Learning for Secure Data in AI Algorithms

Traditional ML algorithms trained on central server datasets face considerable hurdles due to rigorous data security and privacy restrictions. These factors make typical AI and ML approaches unfeasible in data-sensitive and mission-critical situations like health, government, finance, etc. FL has garnered notice for its capacity to protect user privacy by not sending raw data to update global model parameters. FL remains prone to privacy and security threats because attackers can utilize shared gradients to steal participant privacy. To solve this problem, several researchers have proposed various privacy preserving and secure FL approaches. Some of the recent state of the art in this direction are discussed below.

Xu et al. [65] have proposed VerifyNet model that ensures security and verifiability of FL algorithms that train on big data. In the proposed research work, a double-masking protocol is applied on the local gradient data of the users during the FL training process for privacy preservation. Furthermore, the verification of the aggregated results at the cloud server is done. The experiment was conducted on the MNIST dataset, which included 60,000 training set examples and 10,000 test set examples. Experiment results show that the proposed model has low communication overhead and efficient data transmission and processing. In a similar work, So et al. [66] proposed a single-server Byzantine-resilient secure aggregation framework for a secured FL. The proposed approach is based on secure model aggregation, verifiable outlier detection, and integrated stochastic quantization to ensure privacy, convergence, and Byzantine resilience at the same time. The experiment was conducted on the MNIST dataset, and the authors used CNN with 6 layers, ReLu activation function, and 2*2 max pooling layer for training. The results show that the proposed model achieves 99% accuracy. In another interesting work, Li et al. [67] proposed a novel Knowledge Federation framework, to preserve to preserve privacy and securing the training data, utilize all the data resources that are scattered across several units of organizations.

### 3.4. Federated Learning for Other Privacy Preservation Concerns

With an increasing attention on big data privacy preservation mechanisms, several FL frameworks have been developed with high anonymity to protect the privacy of identity information during aggregation stages. For instance, Bin et al. [68] studied an efficient FL approach for anonymous and privacy preservation when mining industrial big data. In this FL framework, besides reducing the communication burden on the FL server when the number of clients increases, the anonymity of all clients was obtained by deploying a proxy server as the middle layer between the server and clients. As a result, the proposed FL framework not only prevents direct and indirect privacy leakage of private data but also guarantees participant anonymity with a middle proxy server to encrypt data. The experiment was conducted on the MNIST dataset, which included 60000 training samples and 10000 test samples. Each image is a handwritten grayscale image with a 28*28 resolution. The results show that the proposed model has an optimal computation cost and a 95% accuracy. In [69], Domingo-Ferrer et al. built a secure and privacy-preserving decentralized FL framework to combat against the Byzantine and poisoning attacks, in which several protocols deployed in the framework can offer strong and sustainable privacy to clients with unlinkable anonymity and co-unitity property. In the framework, a decentralized co-utile reputation management system, including model manager, update generator, update forward, and accountability manager components, was deployed to prompt all clients to adhere to different protocols (such as update submission, reward, and punish). In [70], Chen et al. introduced an anonymous FL framework for IoT systems (called FedTor) by taking advantage of the onion router in dealing with desperate demand of IoT devices for privacy preservation. The FedTor framework achieved high anonymity by hiding user identity in the local model transmission and aggregation stages by adopting multi-layer encryption by maintaining accuracy. In FedTor, there was a reputation-based router selection scheme that realized a lower malicious rate of an onion router without any compromise of transmission delay.

Some privacy-preserving ML models have been proposed to train models with encrypted data on the cloud server in centralized systems, while clients should encrypt and decrypt the private data using homomorphic encryption. In [71], Jia et al. solved the security vulnerability problems (such as model extraction attack and model reverse attack) in FL-based IIoT systems by manipulating multiple data protection schemes in data sharing and model sharing. For data sharing, a distributed K-means clustering scheme was learned with differential privacy and homomorphic encryption. Besides, the authors deployed a distributed random forest with differential privacy and a distributed AdaBoost with homomorphic encryption for model sharing. The experiment was carried out on a Spark distributed cluster with the Adult dataset from the UCI library. The results of the experiment show that the proposed model outperforms other existing models. Zhang et al. [72] introduced DetectPMFL, a privacy-preserving momentum FL framework, with CKKS homomorphic encryption to protect the private information of clients in both local training and global model aggregation stages. For evaluating the credibility of each industrial agent in a FL network, the authors additionally designed a novel method to detect unreliable agents based on CKKS encryption and cosine similarity. Relying on extensive security analysis and experimental results, DetectPMFL achieved high accuracy of FL-based classification tasks while effectively preserving privacy with verified reliable clients.

### *3.5. Summary*

Due to the rapid growth of the applications, large volumes of data are being generated at an exponential rate, encouraging the applications to generate the data at a rapid pace. From the discussion above, it can be summarized that FL has a tremendous potential in addressing several challenges faced by big data services such as big data acquisition, big data storage, big data analytics, big data privacy preservation. However, several companies prefer not to share their data for a variety of reasons and concerns, such data privacy, data sharing rules, and a lack of information that can be solved using FL. Performing big data analytics in the cloud has concerns related to network cost and data privacy. Poor latency for data exchange among decentralized multiple parties, leaking of sensitive information of the participants to the untrusted servers are to be addressed to realize the full potential of FL for big services. Table 4 summarizes the benefits and challenges of FL for big data services.

## 4. Federated Learning enabled Big Data Applications

Federated learning is critical for privacy-sensitive data applications using distributed training data. It seems like a great idea to tackle difficulties with typical machine learning models that store data on one server. FL's uses are diverse, and several are explored as part of this study. In this section, the advantages of FL for several big data applications are discussed along with recent state of the art literature. Figure. 4 shows Federated learning enabled big data in four primary application sectors: smart city, healthcare, smart grid, and smart transportation.

### *4.1. Federated Learning Enabled Big Data in Smart City*

A smart city improves the quality of life of the urban citizens by optimizing the functions of the cities. It also promotes economic growth by making use of data analysis and smart technologies like IoT.

IoT sensors are placed throughout the smart cities to collect live and streaming data pertinent to traffic, air quality, condition of the roads, garbage, etc [76]. Hence there exists the necessity of a scalable, robust and secured framework such as Apache Flink, Apache Spark, and InfluxDB that ensures efficient processing of big data in smart city setups [77] [3]. Big data analytics along with ML-based algorithms can effectively use these data to gain relevant insights. The insights gained can be used to manage the services, resources, and assets in the cities. Big data analytics can provide solutions to some of the important services like healthcare, IoT, communication, transportation, garbage collection, disaster management, etc in smart cities [78]. In a smart city based application, the data from the sensors are generated continuously, in the form of streaming data being acquired at frequent time intervals. Also, most of the data generated from the sensors in smart cities are sensitive as it includes private data of its citizens [79, 80, 81]. Hence, the need of the hour is to process the data at their source to avoid latency issues enable real-time decisions, preserve the privacy of the data generated in smart cities and efficient handling network resources to handle the big data. FL has the potential to play a crucial role in mitigating and addressing the aforementioned challenges of big data in smart cities. Recent studies that have used FL for addressing the challenges of big data from smart cities are discussed below.

[3] https://www.influxdata.com/

Table 4: Benefits and challenges of FL for big data services.

| Services | References | Existing Challenges Faced | Benefits of using FL |
|---|---|---|---|
| Big Data Acquisition | [44, 45, 46, 47, 48, 49, 54, 55, 57] | - Most of the companies prefer not to share their data for a variety of reasons and concerns, such data privacy, data sharing rules, and a lack of information.<br>- Data security and privacy for maintaining the storage of local data.<br>- Difficult to understand and analyse data patterns of fraud transactions and later detect actual frauds.<br>- Poor latency time for data exchange among decentralized multiple parties. | - Training data is held on the device, with no individual updates sent to the cloud.<br>- To understand and learn the global model without compromising with the data privacy.<br>- Achieve data privacy without sharing the data or sensitive information of the users.<br>- Achieve secure and fast data exchange among decentralized multiple parties.<br>- FL with Blockchain build immutable data blocks which provides transparency and improves data ownership without the need of any central authority. |
| Big Data Storage | [49, 16, 58, 54, 55, 56, 57] | - The central server becomes unavailable, the entire training process will be disrupted.<br>- The server have poor bandwidth and, unreliable communication links with the agents to transfer big data.<br>- Poor data storage infrastructure.<br>- Poor latency time for data exchange among decentralized multiple parties. | - FL provides a decentralized approach, where the model relies on the own resources of the node, can be used as an alternative to perform the training.<br>- FL allows mobile agents to collaborate in the models training that does not rely on any central server.<br>- Maintains the heterogeneity of the data, hence reducing the deviation of training in ML model.<br>- Decouples the ability to implement machine learning based on the need to store data in the cloud. |
| Big Data Analytics | [59, 60, 59, 61, 63, 64, 65, 66, 67] | - Big data analytics require the participant sources to send their personal and sensitive data to a central sever to perform ML services,that may result in privacy concerns of the data sources such as mobile users.<br>- Secure Data Training.<br>- Performing big data analytics in the cloud has concerns related to network cost and data privacy.<br>- Leaking of sensitive information of the participants to the untrusted servers.<br>- Conventional AI and ML based approaches are becoming impractical in several data-sensitive and mission critical scenarios, like health, government, finance sectors,etc. | - FL provides privacy reservation while providing big data analysis.<br>- Addresses network connectivity issues and limited computational resources for performing big data analytics in IoT devices that slows down the training process.<br>- Fastening up of the training process of big data from IoT devices.<br>- The network members can download the aggregated model that can be used to provide edge intelligence to the users.<br>- Ensures security and verifiability of FL algorithms that train on big data. |
| Big Data Privacy Preservation | [17, 73, 74, 75] | - Privacy Preservation in Big Data Processing.<br>- Privacy Preservation in Big Data Storage.<br>- Preserving privacy of sensitive information of users.<br>- Poor security and slow data exchange among decentralized multiple parties. | - Solve problems associated with data islands, breaking of data barriers.<br>- Ensures protection of data privacy and security of big data.<br>- Keeps the raw data distributed in the client devices instead of aggregating the same to the centralized cloud data center to perform training.<br>- The shared model is trained on the server by the aggregating the computed updates present locally. |

Prediction of traffic flow plays a significant role in the management of intelligent transportation in smart cities. Analysis of the patterns from the big data related to the traffic flow and then making relevant predictions can help the realization of several benefits like reducing traffic congestion, accidents, and air pollution. However, as the driver's information is tightly coupled with the location of the vehicle, the data collected from the vehicles may threaten the privacy of the driver. The attacker/malicious user can get hold of personal details of the driver like health condition, habits, religion, income, etc. based on the places visited frequently by the driver. Motivated by these issues, Qolomany et al. [82] have proposed to use a decentralized FL system to preserve the privacy of the user, in which a deep LSTM model was developed to train the feature sets of different clients locally (e.g., features are calculated on the same space for different samples). The parameters for the LSTM model are chosen by the particle swarm optimization (PSO) algorithm. Each client computes and shares the update with the global model residing at the server. Thus the private and sensitive data does not get transferred to the central storage thereby preserving the privacy of the drivers. The number of particles considered in this study for the PSO is 5. The root mean square error (RMSE) and accuracy achieved is 95% each. The results obtained indicated the fact that the number of communication rounds in client-server is reduced to the order of 2% to 4%.

The safety monitoring system is a very important application (like monitoring of fire hazards) in a smart city. Building models for object detection/recognition on large datasets that are stored centrally is a very challenging task due to the high cost involved in transmitting video/image data and privacy issues. To overcome these challenges, Liu et al. [83] proposed a machine learning platform, namely, Fedvision, that leverages HFL (a.k.a., sample-based FL) in computer vision-related applications. Fedvision is deployed with the collaboration between Extreme Vision and WeBank for smart city applications that can be used by the customers in developing safety monitoring systems. To train the ML model, a minimum of 1000 sample images are required. To label, the images 5 labellers are required for 2 weeks, which includes testing and packaging, i.e, approximately 30 days is required for model optimization. However, with FedVision, the system administrator can label the images by themselves and the total time for model optimization is reduced by more than 20 days, which saves the labour cost. The proposed FedVision reduced the network bandwidth requirement to less than 1 megabytes/second, which is significantly less when compared to 50 megabytes/second in the conventional ML model. One of the challenges in smart city applications is to process and get insights from large volumes of unlabelled data being generated at a very rapid rate. Also, the privacy of the data has to be preserved. To overcome these challenges, Albaseer et al. [84] proposed an FL based semi-supervised method for real-time analytics at the edge networks, FedSem, for smart city applications. The authors have used the German Traffic Sign dataset to simulate traffic signs in the real world and evaluation of the proposed model. The number of rounds used in the proposed work is 30, 50, and 100 respectively and for each round, the number of clients considered is 10 and 20. The results obtained showed that the proposed FedSem model improved the accuracy by 8%.

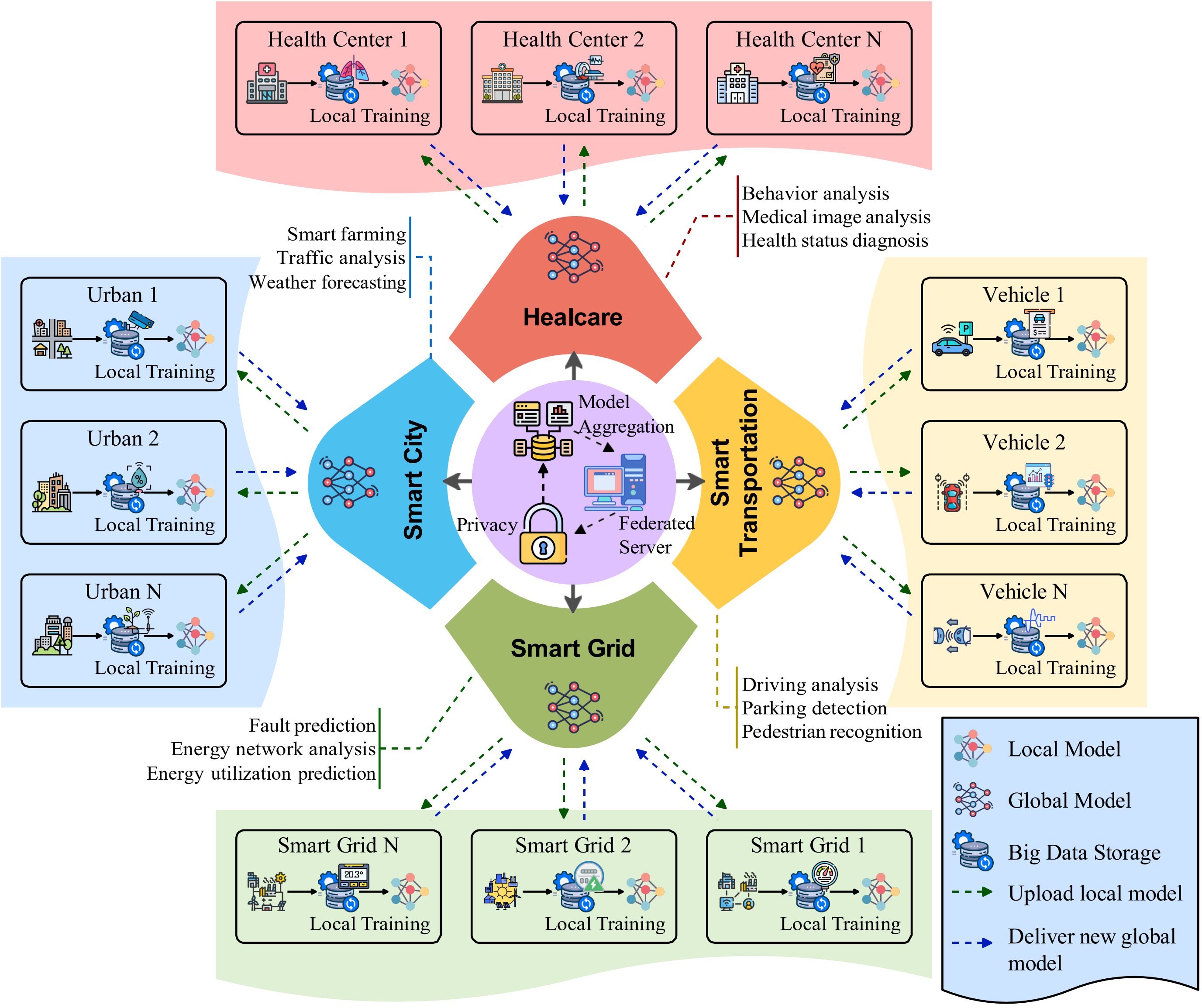

Figure 4: Federated learning enabled big data in four primary application sectors: smart city, healthcare, smart grid, and smart transportation.

One of the primary goals of a smart city is to build a smarter and better healthcare infrastructure. Training large volumes of medical data in a central server involve several challenges such as preservation of the privacy of patients, heterogeneous nature of data from different sources, etc. To address these challenges in large volumes of medical data, Thwal et al. [85] proposed an FL paradigm based on deep learning. In the proposed work, FL was executed 30 times. The proposed model ensures privacy preservation of patients' data and also enables data mining on a large scale.

When faced with a crisis like COVID-19, a digital twin enabled smart city can face several challenges due to its reliance on high quality and long term data. To address these challenges, Pang et al. [86] proposed an FL based city digital twin to efficiently accumulate insights for heterogeneous sources of data. The global model is trained iteratively at different city digital twins till the global model uncovers the patterns between the response plans and trends in infection. In this way, the proposed model can be used to manage the crisis in a city. The authors have used FedAvg as the aggregation algorithm to save the communication resources. The prediction loss of FedAvg is very less for all the cities considered in the proposed study.

Unmanned Aerial Vehicles (UAVs) play an important role in smart cities to monitor traffic, air pollution, movement of citizens during pandemics, and disaster management. However, the data acquired/processed by the UAVs may be very sensitive and the privacy preservation of the data acquired is of paramount importance. Hence, UAV integrated with FL can preserve the privacy of the sensitive data, and address several concerns like resource management, and latency issues faced by UAVs, especially when the terrestrial computing infrastructure is not available [87]. The authors in [88] leveraged FL for UAVs in smart city monitoring for truthful reporting from UAVs and also to address the heterogeneity of data generated from multiple UAVs.

### 4.2. *Federated Learning enabled Big Data in Smart Healthcare*

AI/ML techniques integrated with IoT have been extensively used over the past two decades in healthcare applications to assist the medical practitioners in diagnosing the diseases, drug discovery, remote patient monitoring, etc [89, 90, 91, 92]. Some of the challenges that have to be addressed due to the big data generated from smart healthcare applications are privacy preservation, ensuring the quality of the data acquired from wearable devices, and enabling a collaborative environment for multiple hospitals to work together without the risk of exposing their private data [93, 94]. FL can enable large-scale precision medicine respecting individual privacy concerns, reducing the possibility of noisy data due to data transmission, and enabling the collaboration between multiple hospitals [95]. The recent works on applications of FL for big data related to smart healthcare are presented in the rest of the subsection.

Hakak et al. [96] have proposed an edge assisted framework with FL for privacy preservation of sensitive healthcare data acquired through wearable devices. Apart from privacy preservation, the proposed framework helps in optimizing cloud resources when dealing with huge volumes of healthcare data generated by millions of wearable devices. In a similar work, Qayyam et al. [97] proposed an edge assisted framework for automatic diagnosis of COVID-19 using clustered FL. The proposed framework can help in performing analytics for the large volumes of ultrasound and x-ray data related to COVID-19 patients at the FL enabled edge. The proposed FL approach achieved an improvement of 16% and 11% in overall F1-scores respectively on X-ray and ultrasound datasets.

Silva et al. [98] have proposed a framework based on FL to train the ML models on databanks that contain brain images. The proposed approach is mainly aimed at privacy preservation of the sensitive information of the patients. The proposed FL framework was applied across 200 folds for each scenario.

Wu et al. [99] proposed a novel system, FedHome, that is based on a centralized FL framework with a horizontal data property (i.e., different personal data with the sample feature space collected from different distributed clients) for in-home monitoring of health. The authors aim to address the existing challenges of privacy preservation and transferring of large volumes of data that exist when using the ML algorithms on the in-home patients' data in a central server. In the proposed work, the authors firstly set the local minibatch size as 10 and the number of training passes for local data of each client as 5, varying the number of clients participating. The proposed FL approach has achieved a test accuracy of 95.87% for balanced data and 95.41% for imbalanced data. To get diverse and large datasets to train a deep learning model, multi-institutional collaboration is the need of the hour to generate a diverse and large quantity of medical data. How to enable collaboration among multiple institutions without exposing the sensitive information of the patients is a challenge. To address this issue, Shellar et al. [100] proposed an FL based deep learning model for privacy preservation in multi-institutional collaborations, where the proposed model is trained at all the participating individual institutions without the need to share the individual data among the institutions. In a similar work, Wang et al. [101] proposed a 5G-enabled architecture for auxiliary diagnosis of COVID-19 patients based on a centralized FL with multimodal data (including basic identification data, medical record text data, biochemical inspection data, and medical image data) of different patients collected by distributed hospitals. In this architecture, multiple institutes can collaborate without compromising privacy concerns. To measure the effectiveness of the proposed model, the authors have conducted the COVID-19 classification experiments on the central cloud and 3 edge servers. The experimental results showed that the proposed FL aggregation model achieved good performance with an accuracy of 95.3%, 79.4%, and 97.7% on the distributed nodes. The authors in [102] have followed a similar approach for multi-institutional collaboration in the screening of COVID-19 patients based on chest X-ray images using convolution neural networks (CNN). The authors have used 4 clients with similar loss functions and CNN architecture. The highest accuracy achieved is 97%, specificity achieved is 98.11%, and specificity of 95.89% is achieved in this work.

Li et al. [103] have proposed an FL based framework for analysing the MRI images. The proposed work aims to reduce the time and communication cost involved in transmit-ting the MRI images from local sites to the central server for storing and training the ML-based models. In this work the authors have performed a 5-fold cross-validation and used cross-entropy as the objective function. The proposed FL based approach yielded a maximum accuracy of 74.4%.

Even though FL can address the privacy concerns and latency issues in transferring the data from the local devices to the central servers to train the ML models, transparency and tampering of the records persist. To address these issues, blockchain can be integrated with FL for identifying the patterns from large volumes of medical data while preserving privacy and ensuring security. In this direction, El Rifai et al. [104] have integrated FL and blockchain to train diabetes, kidney, and digestive datasets. FL is used to ensure that privacy preservation is accomplished for sensitive data of patients and also to reduce the communication overhead involved in frequently transferring the patients' data from local devices to the central server, whereas blockchain is used to ensure transparency and prevent alterations of the medical records. In this work, the authors have set up the FL setting by distributing the medical records across 15 different collaborative clients. The results obtained show that the accuracy of individual models is lower than that of the centralized setting with an average accuracy of around 73%. All the local models are benefited from the FL setting and their accuracy is improved from 1% to 5%. In a similar work, the authors in [105] proposed a blockchain-aided decentralized FL framework to detect the COVID-19 from CT images, in which the image data is collected from different hospitals and devices (e.g., CT scanner machines having different imaging features). The data is authenticated by blockchain technology, whereas FL trains the model globally thereby achieving privacy preservation. In this work, the datasets were distributed across 3 hospitals for training the FL. The proposed FL based approach outperformed the other existing state-of-the-art by achieving the highest accuracy of 98.68% accuracy and 98% sensitivity.

### *4.3. Federated Learning enabled Big Data in Smart Transportation*

Some of the applications of smart transport systems such as intelligent transportation systems and cooperative autonomous vehicles involve sensitive data and a large number of devices where the storage, computing, and communication resources must be utilized efficiently [106, 107]. Also, the decisions should be made based on the local scenarios rather than the global scenarios to take appropriate actions in specific locations [108, 109]. To address the aforementioned issues, FL can be a promising solution. FL can be used to train the data from vehicles in a particular area collected and stored at local devices in near proximity to analyze and extract the patterns from the local area to take immediate actions, preserving the privacy related to sensitive information of the vehicles such as driver details, location of the vehicle, etc. [110]. Road surveillance is an essential application of ITS, where a large number of videos generated at a fast pace have to be analyzed in real-time for the appropriate authorities to take necessary action regarding tasks such as traffic congestion prevention, immediate response to accidents, burglaries on highways, etc. Even though edge computing is promising, new challenges emerge with it like fragmenting the decision models and coordination between the local nodes distributed in different locations [111]. To address the issues of edge-based frameworks when dealing with real-time video analytics of ITS, Sada et al. [112] presented an FL based architecture for video analytics which is distributed. The proposed architecture enables distributed and real-time object detection apart from preserving the privacy of the model updates. In a similar work, Xu et al. [113] proposed an FL based traffic monitoring system to obtain the data related to the traffic on roads and other locations using remote sensing data for identifying traffic congestion. FL helps in preserving the privacy of sensitive data related to the vehicles, drivers, etc. The authors experimented on remote sensing image datasets from Washington Road and Los Angeles Road to train the proposed model. An accuracy of 85% is obtained for the proposed model. The processing time of each image is approximately 0.047 seconds. In another work, Qi et al. [114] proposed a framework based on FL integrated with blockchain for traffic flow prediction. Even though FL helps in preserving privacy, it has some serious security issues such as the single point of failure of a centralized model coordinator. To address this issue, the authors have proposed a consortium blockchain-based approach that can provide reliable, secure and decentralized FL without the need for a centralized model coordinator. In this work, the miners in the blockchain verify the updates from the models of decentralized vehicles which helps in preventing unreliable model updates. The authors have set the number of vehicles as 10, learning rate as 0.001, number of rounds for training as 500, batch size as 128 and number of training epochs as 5. The results obtained proved that the mean absolute error of the proposed FL based model is 10.04% lower than that of the conventional SVM algorithm with improved privacy and security against poisoning attacks. Similarly, Liu et al. [115] proposed FedGRU, a centralized FL-based gated recurrent unit neural network algorithm to predict the flow of traffic for protecting the privacy of large volumes of sensitive data generated from the vehicles. In this work, the authors have distributed the dataset across 20 organizations for FL setting. A learning rate of 0.001, and a batch size of 128 are the parameters used for training the proposed FL model. The proposed FedGRU model achieved a mean absolute error value of 6.85 when the number of clients is 10 which is better when compared to other state-of-the-art aggregation schemes. Mowla et al. [116] proposed a centralized HFL framework for detecting on-device jamming attacks in flying ad-hoc networks, in which the meaningful information of packet delivery rate and received signal strength indicator are computed on the sample feature space for different samples at distributed nodes in the network. The proposed approach helps in reducing the power consumption, and communication overhead issues faced by traditional jamming attack detection mechanisms, along with providing on-time global updates in flying ad-hoc networks. The proposed approach attained an accuracy of 89.73% for the FANET dataset which is simulated by ns-3 and 82.01% for the CRAWDAD dataset.

In electrical vehicular networks, prediction of the energy demand is of paramount significance as it ensures accurate charge planning and reliable routing. Saputra et al. [117] proposed an FL based architecture to find the complex patterns in the electrical vehicular networks to improve the accuracy of energy demand prediction. To protect the privacy of sensitive data that will be frequently exchanged between the charging stations and the vehicles, and also to reduce the communication overhead, the authors have used FL which allows the charging stations to share the data without exposing the sensitive data. The experimental results prove that the accuracy is improved by 24.63% for predicting the energy demand and a reduction of 83.4% in communication overhead is achieved by the proposed FL based approach when compared with baseline ML algorithms. Similarly, the authors in [118] studied a privacy-preserving and communication-efficient method to predict energy demand and driving range for electric vehicles by exploiting a centralized FL framework. Concretely, the regular FL averaging algorithm was improved to address the predictive uncertainty of traditional FL based models, in which two different feature spaces (the first one includes vehicle speed, road slope, curvature, traffic phase, and traffic speed, and the second extends with segment speed difference, strength, and standard deviation) are formulated for learning regression models and probabilistic neural networks. The experimental analysis proved that the energy demand prediction of the proposed approach is improved by 24.63% and the communication overhead is reduced by 83.4% when compared with baseline ML algorithms.

### *4.4. Federated Learning enabled Big Data in Smart Grid*

The electric grid concept is ageing and the current electric grid infrastructure is finding it difficult to handle the things it is not designed for. Using cutting edge technologies like IoT, ML, communication networks, and equipment in electric grids will ensure that the electricity is delivered more efficiently and reliably. The modernized electric grids, termed smart grids, will result in reducing the frequent and long power outages, faster restoration of services when the outages occur, reduce the impacts of the storms on the electricity supply, etc.

Smart grids generate a large quantity of data at a very rapid pace. ML algorithms can be used on this data to analyze the patterns and predict the electricity demand, power outages, stability of the smart grid, electricity theft, etc [119, 120]. FL can address several issues faced by traditional ML algorithms in handling big data generated from smart grids such as privacy preservation of sensitive data from consumers, appliances, smart grid operators, etc., increased latency and communication costs. The recent literature related to the application of smart FL for smart grids is discussed in the rest of the subsection.

Deep learning models can expose privacy-sensitive data such as the data related to the appliances collected by a smart meter, which may reveal the consumer's behaviour at home based on the appliances used by them, which can raise serious security concerns. Also, the ML/deep learning models need a large quantity of data to get trained. To address these issues, Taik et al. [121] proposed a model based on centralized FL and edge computing for household load forecasting that addresses the volume and diversity requirements of the deep learning models and also preserves the privacy of consumer data. The authors have used the data from 200 houses in Texas, USA for the FL setting. The experimental results show that the proposed FL based model achieves a maximum of 5.07% improvement in mean absolute percentage error and a maximum gain of 97% in network load. In a similar work, Wang et al. [63] proposed an FL based method to identify the electricity characteristics of the distributed consumers that can preserve the privacy of the retailers. In the proposed work, the authors have distributed the training dataset to 5 retailers, where the number of consumers in each retailer is approximately 16%. The proposed FL based approach achieved an improvement of 1.69% in accuracy and 12.5% in Matthews correlation coefficient when compared to centralized ML models.

The future energy demand of a smart home can be predicted by training a deep learning model on the energy consumption data of a consumer. The data collected from the clients who are distributed for centralized model training incurs huge costs concerning communication resources. To address this issue, Tun et al. [122] proposed an FL based model, in which every client has to upload only the model updates generated by training the FL model on its local data and these updates are then aggregated into a global model at the centralized server. The authors have chosen 15 houses for training the FL model. The experimental results proved that the models based on clusters yielded optimal accuracy when compared with a single FL model trained with all the clients.

Forecasting a load of a power system in short term plays an important role in the efficient, stable management and dispatch of the electricity. The traditional model of centralized training of the ML-based models for forecasting the load of the power system has issues regarding privacy preservation and also the latency incurred due to the large volumes of data generated from distributed consumers. To address these issues for electric load prediction, Li et al. [123] proposed a centralized HFL based approach for short-term load forecasting of the grid. In this work, for the FL setting, 2, 4, and 6 edge nodes are set and every node is executed for 5, 10, and 15 rounds respectively for global training of the FL model. The accuracy attained by the proposed FL based model is more than 90%. In a similar work, Savi et al. [124] proposed a framework based on edge-computing and decentralized FL that use LSTM models for short-term consumption of energy of the consumers. In this FL framework, many distributed nodes collected several weather- and calendar-related features clustered in the same space to train LSTM models locally. In the proposed FL based model, the data exchanged is 50 times lesser per hour when compared to baseline models. The proposed FL model attained almost 25% better forecasting performance when compared to baseline models.

### *4.5. Federated Learning enabled Big Data for Online Recommender Systems*

Recommender systems are seen commonly these days in several online applications such as amazon, youtube, Netflix, etc. Recommender systems suggest items that are relevant to the users, such as products to buy, text to read, movies to watch, songs to listen to, etc. Typically. the recommender systems have to process large volumes of data related to the activities of the users to filter, rank and suggest/recommend the items based on their interests [125]. Users browse a large volume of items daily. In recommender systems, a large quantity of text and browsing history is generated frequently that has to be processed at regular intervals. The preferences/recommendations have to be customized at region/individual levels due to the differences in choices based on a demographic or individual basis. FL can play a vital role in preserving the privacy of the users, giving personalized recommendations to the users in near real-time by analysing the data at their source of an edge device that is in proximity to the data source. Some of the recent state-of-the-art applications of FL for recommender systems that have to deal with big data are discussed below.

Amir et al. [126] proposed an FL-based model to improve the personalized recommendations to the users in a privacy-preserving manner. The proposed system ensures that the ratings of the users on the items remain on their local device, and only the preference patterns are shared to the global model which reduces communication overhead. The data is distributed across 943 edge devices for the FL setting. 10 global epochs are executed by the edge devices, after which, 100 local epochs are executed. The experimental results prove that the proposed FL based approach reaches the root mean square value of 1.06 in 80 epochs lesser than the baseline ML algorithms. In a similar work, Tan et al. [127] proposed an FL based approach that trains the data from several clients, while at the same time, preserving the privacy of each client. The authors have implemented the proposed FL based framework in a real-world application. The experimental results proved that the average reading time is improved by 22% and click-through size is improved by 11% when the proposed FL based model is used.

In another interesting work, the authors in [128] developed a secure recommender system with centralized FL to preserve the privacy of the user's preferences, in which a local model of a certain client was trained with a small subspace data instead of the full feature space. The users have tested their proposed approach on Alibaba's e-commerce recommendation and evaluated their model on Taobao user data collected over 30 days. The results proved that the proposed system's scalability, reduced communication, storage and computational overhead while preserving the privacy of the user's data simultaneously. Muhammed et al. [129] presented a novel FL approach in which, a diverse set of sample data from different clients is trained on the federated model and the model updates are propagated to the other clients using an active aggregation technique. In this way, the users can benefit from the proposed approach with reduced communication costs and also the models which are more accurate and can be consumed by the users at even very early stages of training.

### 4.6. Federated Learning enabled Big Data for 5G and 6G

Due to the advent of fifth-generation (5G) cellular networks, the number of wireless devices will increase stupendously reaching the scale of trillions by the next decade. The advantages of 5G include enhanced coverage, and transmission bandwidth along with the reduction in communication latency in versatile applications of smart homes, autonomous vehicles, UAVs and various others. The average 5G speed will soon be much higher than the average mobile connectivity and this has resulted in the announcement of over 81 billion dollars bidding for 280MHz licenses of the aforementioned frequency. To use this valuable spectrum resource, the third generation partnership projects (3GPP) have been trying to standardize and enable 5G new radio services in unlicensed spectrum being part of LTE which is a Licensed Assisted Access initiative (LAA). FL acts as a hybrid technique involving centralized and decentralized learning. The model training is done using the decentralized technique and the data gets stored at its source. The raw data is stopped from being transferred to prevent communication overhead thereby optimizing performance and security [130, 131, 132]. The study in [133] proposed a fog computing and FL - based network architecture wherein a large number of devices used for various services were integrated to construct a common model for image classification, voice recognition, next-word prediction and anomaly detection. The study in [134] proposed a distributed federated learning framework (DBFL) that helped to mitigate energy efficiency-related issues in case of communicating with distant devices. The edge computing techniques helped to connect the devices using different clustering protocols. Notably, DBFL can facilitate homogeneous (i.e., the training data at all local devices should have the same feature space) and heterogeneous (i.e., devices may train the local models on different feature sets extracted by different feature extraction techniques) feature spaces, which is applicable to various networking scenarios. The proposed FL based framework improves the performance of the classification by approximately 7.4% when compared to conventional FL, reducing the consumption of energy simultaneously.

The study in [135] proposed a multi-layer heterogeneous model selection and aggregation scheme incorporated with centralized FL that enabled enhanced utilization of local and global contexts of vehicles and roadside units (RSUs) in 6G vehicular network systems. All connected vehicles in a network should train the local models on their datasets with identical features, and then submit the local training parameters to aggregate them and calculate the global parameters at RSUs and the central cloud server, respectively. The results obtained proved that the proposed FL based model outperforms the baseline ML model by achieving a precision of 90.6%, recall of 96.8%, and F1 score of 93.6%.

### 4.7. Federated Learning enabled Big Data for Industry 4.0/5.0

FL has evolved as an extremely successful solution for the development of cost-effective and secured IIoT applications. The implementation of FL helps in attracting large datasets and computation resources from various IIoT devices for the training of the AI models. This improves the quality of IIoT training data which may not be possible using the traditional AI approaches. The integration of FL with IIoT has various advantages such as privacy preservation, and reduced communication costs. This motivates the use of FL in applications for mo-bile devices, manufacturing robots, industrial automation, etc [20, 136].

### *4.8. Federated Learning enabled Big Data for Social Media*

These days most individuals are spending a great amount of their time on their mobile phones, especially, on social media. In several social media apps such as Facebook, Instagram, Twitter, etc., the users will be posting photos, and videos, tweeting their status, sharing their opinions, sharing other posts, etc., which results in the generation of heterogeneous data at a very fast rate and of enormous proportions [137]. Conventional big data analytics may incur several issues such as leakage of sensitive information of social media users such as their preferences, location, etc. To address these issues, FL can be utilized to train ML models on the user's mobile devices, which would result in privacy preservation of the social media users, optimize the resources, and improve the real-time analytics of the big data related to social media [138].

### *4.9. Federated Learning enabled Big Data for Other Applications*

FL can be used for several other applications that are based on big data such as pollution control, defence etc. Liu et al. [131] proposed a novel FL based framework for monitoring and predicting the quality of the air. The proposed framework predicts the air quality index scale distribution by using a Dense-MobileNet model that learns the haze features from the images taken by UAVs in the air. The traditional method of transferring the large volumes of the data generated by the UAVs incurs heavy communication overheads. Also, the generated data may include sensitive information that should not be exposed. The proposed FL enabled architecture ensures several organizations learn collaboratively through a global model to monitor the air quality index in a privacy-preserving manner. The authors proposed a long short-term memory model based on graph convolutional neural networks for ground sensing systems to achieve real-time, accurate and future air quality index inference. FL enabled technologies can be effectively used by UAVs in several big data applications such as predicting optimal routing path, forecasting the mobility patterns on the ground users, monitoring in crisis/disaster situations, defence etc [139, 140]. FL enabled big data frameworks can also be used in several applications such as online shopping, insurance, credit card divisions etc., where the privacy of the personal information of the users has to be preserved.

### *4.10. Summary*

It can thus be summarized that FL has great potential to solve big data application challenges like privacy preservation, resource management to handle a large volume of data, communication cost and latency issues in transferring large volumes of data at a rapid pace to the central server for ML algorithm training, real-time analytics, customized decisions based on geographical locations, and heterogeneity. To fully utilize FL in big data applications, various issues must be addressed, including the poisoning of ML models in local devices, malicious terminals that may mistrain FL models, and excess energy requirements at local devices. Table 5, 6, 7 summarizes the benefits and challenges of FL in big data applications.

## 5. Federated Learning Big Data Projects

Studies and researches pertaining to FL implementations are still at a nascent stages. However, some projects have achieved promising outcomes. This section presents some of the such FL projects and platforms which are widely used for Big Data developments.

### *5.1. Flower: A Friendly Federated Learning Research Framework*

Flower is an open-source platform-independent FL framework[4]. Flower framework is specifically designed to provide scalability, which is appropriate to handle big data applications. Flower framework has been used for many real-world applications, which can also support more than 10.000 clients. Moreover, Flower framework supports a wide range of devices, including Android mobiles, iOS devices, Raspberry Pi and Nvidia Jetson.

Key features of Flower framework are as follows[141].

- **Support for heterogeneous clients**: Flower can offer the same workload to clients running on different operating systems using different languages.
- **Provide High Scalability**: by allowing the workload to scale across thousands of machines.
- **Separation of Client-side and Server-side definition** which allows independent control over local client-side and global server-side computations.
- **Support of ML framework-agnostic libraries:** allows the users to use their favourite and experienced ML frameworks
- **Language-agnostic support:** allows users to implement FL clients in different languages and emerging embedded device platforms
- **Support for baselines:** allows fast comparison of the new FL algorithms with existing and well-known algorithms.

### *5.2. Pysyft*

PySyft is one of the most popular open-source deep learning frameworks which offer privacy and support FL[5]. It is a Python library to perform secure and private deep learning methods[142]. Moreover, PySyft can be integrated with other deep learning frameworks such as PyTorch, Keras, or TensorFlow.

[4]https://flower.dev/

[5]https://blog.openmined.org/tag/pysyft/

Table 5: Benefits and challenges of FL for big data applications.

| Applications | Ref.No. | Big Data Acquisition | Big Data Storage | Big Data Analytics | Big Data Privacy Preservation | FL Analytics Algorithms/computing Platform | Privacy-Preserving Techniques | Remarks |
|---|---|---|---|---|---|---|---|---|
| Smart City | [82] | ✓ | ✓ | ✓ | × | PSO based methodology for training local ML model. | No specific privacy preservation technique was applied. | Fine-tuning of the DL parameters and FL associated parameters in deciding the specific number of clients in a particular cluster is not included. Integrating PSO with DL enhances client-server communication to find optimal parameter settings. |
| | [83] | ✓ | ✓ | ✓ | × | FedVision Framework Visual Object detection algorithm for training local devices | No specific privacy preservation technique was applied. | Due to lack of explainability, trust and reliability issues exist. In future, integration of XAI and Incentive mechanism will assist in developing sustainable FL eco- system. |
| | [84] | ✓ | ✓ | ✓ | × | FedSem Framework(Semisupervised Federated Edge Learning) | - | In the case of labeled data accuracy could be improved. In the future, more experiments can be conducted by integrating XAI methodologies. |
| | [85] | ✓ | ✓ | ✓ | ✓ | Customized Federated Learning Framework | Randomization | The communication cost of the federated model is relatively high due to the noisiness of data. |
| | [86] | ✓ | ✓ | ✓ | ✓ | Local Digitaltwin Framework | GRU based encoder-decoder structure | The applicability of the proposed framework could be tested with more data sources to further justify the performance |
| | [88] | ✓ | ✓ | ✓ | ✓ | FL with Gale-Shapley algorithm | Privacy-preserving collaborative training model for the Internet of Vehicles in intelligent transport systems. | Inclusion of Wireless charging Techniques could be incorporated to ensure simultaneous sensing and model training |
| Smart Healthcare | [96] | ✓ | ✓ | ✓ | ✓ | Age-assistive data analytics based FL-framework | FedAvg as part of cloud module | Vulnerability of edge devices to DoS attacks and data biasing issues |
| | [97] | ✓ | ✓ | ✓ | ✓ | Clustered FL integrated VGG16 model | - | Variance in data distribution considering a high number of samples may be considered in the future. |
| | [98] | ✓ | ✓ | ✓ | ✓ | Federated Principle Component Analysis) | Data distribution | Implementation of the model on large-scale imaging genetics data could be included. |
| | [99] | ✓ | ✓ | ✓ | ✓ | FedHome (Home monitoring of health) | Data distribution | Implementation of the model on a larger dataset can be included |

PySyft framework is leveraging three main techniques, i.e., Secured Multi-Party Computations (sMPC), differential privacy, and FL. The use of sMPC in PySyft helps to protect the privacy of input data. sMPC can be used to perform computations over private input data. Differential privacy can reduce the statistical privacy leakages in large datasets. In addition, PySyft also supports Homomorphic Encryption (HE) to offer extra privacy.

### *5.3. Fate: An Industrial Grade Federated Learning Framework*

Fate is another open-source FL framework developed by Webank's AI department[6]. FE Fate platform primarily supports big data collaboration according to the regulations. It was enabled by integrating multiple secure computation protocols in the Fate framework. Fate platform uses several features such as a flexible scheduling system, a modular, scalable modelling pipeline, and clear visual interfaces to keep the scalability, user-friendliness, and improved operational performance.

[6]https://fate.fedai.org/

Table 6: Benefits and challenges of FL for big data applications *(Continued)*.

| Applications | Ref.No. | Big Data Acquisition | Big Data Storage | Big Data Analytics | Big Data Privacy Preservation | FL Analytics Algorithms/computing Platform | Privacy-Preserving Techniques | Remarks |
|---|---|---|---|---|---|---|---|---|
| Smart Transportation | [112] | ✓ | ✓ | ✓ | ✓ | FL Integrated Video Analytics using CNN | Data distribution | Many challenges related to the inner FL architecture need to be addressed and implemented. |
| | [113] | ✓ | ✓ | ✓ | ✓ | FL based traffic monitoring system using CNN | - | During the data transmission process, the asymmetric encryption algorithm is used. Implementation of congestion detection techniques using enhanced comprehensive remote sensing image and video streams could be included in the future. |
| | [114] | ✓ | ✓ | ✓ | ✓ | Consortium blockchain-based FL integrated Framework | Differential privacy method integrated with noise adding mechanism | The proposed method can provide protection against unreliable traffic flow updates and predictions ensuring security and privacy. |
| | [116] | ✓ | | ✓ | ✓ | FL based model for detecting on-device jamming attacks | Damster-shafer method | The reliability of the global updates could be enhanced using decentralized model in the UAV-fanet architecture |
| Smart Grid | [121] | ✓ | ✓ | ✓ | ✓ | Edge computing-enabled Federated Learning using a decentralized machine learning model | Secure Aggregation and differential privacy method | Integration of Neural networks with clustering or similar methodologies could be implemented to enhance the accuracy. Also, outlier issues could be further investigated. |
| | [122] | ✓ | ✓ | ✓ | ✓ | Recurrent Neural Network-based Energy demand predictor integrated with FL | Data distribution | More experiments can be carried out on diversified datasets. |
| | [123] | ✓ | ✓ | ✓ | ✓ | Fl integrated Ultra short-term load forecasting method for implementing power Internet of Things. | - | The problem of fairness and privacy can be addressed in the future. |
| | [124] | ✓ | ✓ | ✓ | ✓ | Edge computing-enabled and FL-integrated LSTM model | Localization of data | Integrating differential privacy principles to avoid inversion attacks could be included. |
| Online Recommendar System | [126] | ✓ | ✓ | ✓ | ✓ | FL based recommender system using meta-learning algorithms | REPTILE meta-learning algorithm | Privacy preservation in FL using asynchronous aggregation schemes could be explored |
| | [127] | ✓ | ✓ | ✓ | ✓ | Federated Recommender System using general matrix factorization, SVD, factorization machine, wide and deep learning algorithm | Secure instance mapping and encryption of model parameters are implemented | Realtime implementation of the FL system with diversified data could be explored |
| | [128] | ✓ | ✓ | ✓ | ✓ | Secured FL based submodel learning scheme | Private set union protocol, randomised response, secure aggregation, Bloom filter | The framework could be implemented in realtime live scenario to validate its efficiency |
| | [129] | ✓ | ✓ | ✓ | ✓ | FL based distributed learning technique | Differential privacy enhancing techniques | Incorporation of reinforcement learning with FL could be considered |

Fate platform supports different service modules such as

- **FederatedML**: supports common FL/ machine learning algorithms and tools such as DataIO, Intersect, Federated

Table 7: Benefits and challenges of FL for big data applications *(Continued)*.

| Applications | Ref.No. | Big Data Acquisition | Big Data Storage | Big Data Analytics | Big Data Privacy Preservation | FL Analytics Algorithms/computing Platform | Privacy-Preserving Techniques | Remarks |
|---|---|---|---|---|---|---|---|---|
| Social Media | [138] | ✓ | ✓ | ✓ | ✓ | FL integrated mobile devices for providing privacy preservation to social media users | Data distribution | Improvement of real-time analytics of social media users and optimization of resources can be done. |
| | [137] | ✓ | ✓ | ✓ | ✓ | FL integrated mobile devices for providing privacy preservation to social media users | Data distribution | Improvement of real-time analytics of social media users and optimization of resources can be done. |
| 5G and 6G | [133] | ✓ | ✓ | ✓ | ✓ | Fog computing and FL-based network architecture | Secured aggregation techniques implemented at the fog servers | Optimization of resources, integration of XAI, granularized privacy control could be incorporated |
| | [134] | ✓ | ✓ | ✓ | ✓ | Edge computing based distributed federated learning framewor | Meta learning and Weighted averaging techniques in Mobile Edge Computing Framework (MECF) | Selection of verification and validation measures abiding to the GDPR regulations need to be considered |
| | [135] | ✓ | ✓ | ✓ | ✓ | FL based multi-layered heterogeneous model selection and aggregation scheme for 6G enabled vehicular networks | Two layered FL architecture for 6G based vehicular networks with end-edge cloud components | Incorporation of deep learning with FL could be explored to achieve enhanced accuracy and efficiency |
| Industry 4.0 — 5. | [20] | ✓ | ✓ | ✓ | ✓ | FL integrated IIoT for privacy preservation | Data distribution | Improvement of the training process of IIoT data is required. |
| | [136] | ✓ | ✓ | ✓ | ✓ | FL integrated IIoT for privacy preservation | Data distribution | Improvement of the training process of IIoT data is required. |
| | [131] | ✓ | × | ✓ | ✓ | FL-integrated framework for air quality monitoring using CNN | Data distribution | The data generated by UAVs generate enormous communication heads. FL enabled big data frameworks can provide optimized results. |

Sampling, Feature Scale, Hetero Feature Binning, One-Hot Encoder, Hetero Feature Selection, Hetero LR, Hetero Poisson, Homo LR, Homo NN, and Hetero Secure Boosting.

- **FATE-Serving**: support high-performance FL algorithms.
- **FATE-Flow**: is an end-to-end pipeline platform to perform highly flexible and high-performance FL tasks.
- **FATEBoard**: is a tool to visualize the FL models. It is useful to explore, improve and debug the FL models efficiently.
- **FATE Network**: is designed to enable secure and efficient communications between FL parties.
- **KubeFATE**: manages the workload of FL tasks to achieve high scalability, high efficiency, and cost reduction.

### *5.4. FedML: Federated Learning Library*

FedML is an open research library and benchmark which supports the deployment of evaluation of novel FL algorithms [143]. FedML supports standalone simulation, distributed training, and mobile on-device training like many other FL platforms. FedML supports worker/client oriented programming to offer an easy user experience. Moreover, it enables end-to-end design pattern toolkits to compare and evaluate novel custom FL algorithms with existing known algorithms.

### *5.5. TensorFlow*

TensorFlow[7] is also widely used open-source machine learning platform. The Google Brain team originally developed it for their internal use. TensorFlow supports a wide range of flexible tools which can be used to deploy ML-powered applications and build customized ML models.

[7] https://www.tensorflow.org/

TensorFlow Federated (TFF) [8] is extension of TensorFlow platform. TFF is specifically designed to support decentralized data processing and FL. TFF contains a rich set of tools that users can use to run FL algorithms on their models and data. Moreover, TFF supports the experimentation of new FL algorithms.

There are two layers in the TFF platform.

- **FL Layer**: It provides the interfaces to interconnect existing Keras or non-Keras machine learning models with the TFF platform. This layer supports the users in conducting basic FL tasks such as FL training and FL evaluation. Here, users do not need to be fully aware of the mechanism or structure of the FL algorithm to run the tests.
- **Federated Core (FC) Layer**: This layer support customizes user FL algorithms. It offers low-level interfaces which can be used to interconnect TensorFlow with distributed communication operators and perform custom FL algorithms.

### *5.6. LEAF: A Benchmark for Federated Settings*

LEAF is a popular modular benchmark framework for FL settings. It supports the benchmark of FL's different applications, including multi-task learning, meta-learning, and on-device learning[9].

LEAF framework includes three main components[144].

- **A suite of open-source datasets**: This is responsible of prepossess the data and covert to standardized fromat. LEAF support different well-know data sets such as Federated Extended MNIST (FEMNIST)[145], Sentiment140[146], Shakespeare[147], CelebA[148], Reddit and a Synthetic dataset [149]
- **An array of statistical and systems metrics**: This contains the different evaluation metrics which can be used to assess the behaviour of learning solutions in federated scenarios.
- **A set of reference implementations**: This contains a set of reference implementations such as Mocha[150], FedAvg[151] and minibatch SGD, which support FL.

Thus, LEAF allows researchers in FL to benchmark their novel and customized FL solutions under realistic settings. The comparison the popular FL platforms is summarized in Table 8.

## 6. Research Challenges and Future Directions

As already discussed, FL is a machine learning-based technique wherein various clients collaboratively perform training of a model under the control of a centralized server but the training data remains decentralized. Machine learning algorithms, particularly deep neural networks get trained using multiple datasets being part of local nodes. The raw data in a traditional system gets aggregated to a centralized data centre for the purpose of training. But in the case of FL, the raw data is left distributed on the client devices being further trained as a shared model on the server through the aggregation of local computed updates. There exist several research challenges associated with FL in big data applications. These challenges can be broadly dichotomized into two categories namely the training related challenges and security-related challenges. The data training related challenges include communication overhead while performing multiple training iterations. It also includes issues relevant to the heterogeneity of the devices used in learning and the heterogeneity of the data used in training. On the other hand, security challenges include threats to privacy and security caused by adversaries starting from malicious clients in the local device to malicious users who have black-box access to a model. In the case of FL, the private data remains inside the device instead of leaving and thus makes it easier for an unauthorized user to learn about the existence of the data point used for the training in the local models. Security attacks are often introduced due to the manifestation of targeted or non-targeted malicious attacks in the learning process. In the case of the targeted attacks, the adversary manipulates the labels of specific tasks. In the case of non-targeted attacks, the attacker primarily intends to compromise the accuracy of the global model [152]. Some of the core challenges in the implementation of FL in big data applications are discussed in the next sections as illustrated in Figure. 5.

### *6.1. Bottleneck in Communication Efficiency*

Communication emerges as a bottleneck in the development of FL based systems. The primary reason for this is the inclusion of a massive number of devices and communications in the network which could be slower than the local computation of other magnitude orders. FL sends small messages or models iteratively being part of the distributed training process instead of transporting the entire dataset over the network. The main communication challenges associated with FL for Big Data primarily revolve around the substantial communication overhead caused by frequent model updates between clients and the central server, resulting in significant bandwidth consumption and latency. As the number of clients increases, scalability issues emerge, which limits effective execution and synchronization. Client heterogeneity leads to inconsistent and unreliable data transmission. Implementing robust encryption protocols and robust security measures to safeguard communication and prevent unauthorized access to sensitive data is a highly complex process. In addition, the challenge lies in synchronizing communication between clients to guarantee consistent updates of the model, while also dealing with the high cost of data transmission in environments with limited resources. Dealing with communication failures and guaranteeing reliable updates for models in the context of unreliable network conditions is challenging. This requires the use of efficient communication protocols and strategies to optimize Federated Learning in Big Data environments. The following strategies act as potential

[8] https://www.tensorflow.org/federated
[9] https://leaf.cmu.edu/

Table 8: Comparison of existing FL platforms [141] [143] [142] [144]

| | | FL Platforms | | | | | |
|---|---|---|---|---|---|---|---|
| | | Flower | Pysyft | Fate | FedML | TFF | LEAF |
| Algorithms | SplitNN | ✓ | | | ✓ | | |
| | FedNAS | ✓ | | | ✓ | | |
| | FedAvg | ✓ | ✓ | ✓ | ✓ | ✓ | ✓ |
| | VFL | ✓ | | ✓ | ✓ | | |
| | decentralized FL | ✓ | | | ✓ | | |
| Computing Platform | On-Device | ✓ | | | ✓ | | |
| | Standalone | ✓ | ✓ | ✓ | ✓ | ✓ | ✓ |
| | Distributed | ✓ | ✓ | ✓ | ✓ | ✓ | |
| Benchmark | vertical FL | ✓ | | ✓ | ✓ | | |
| | Shallow NN | ✓ | ✓ | ✓ | ✓ | ✓ | ✓ |
| | Model DNN | ✓ | | | ✓ | | |
| | Linear models | ✓ | ✓ | ✓ | ✓ | ✓ | ✓ |
| API Design | Customizable typologies and message | ✓ | ✓ | | ✓ | | |
| | Message flow flexibility | ✓ | | | ✓ | | |

concepts to achieve communication efficient learning methods [153, 152, 154].

**Possible Solution and Direction of Future Research**
The first strategy is to use Local updating methods that allow the variable number of local updates to be implemented on each machine in parallel during each communication round. This would help lessen the number of communication rounds. Secondly, model compression mechanisms namely sparsification, subsampling and quantizations also help to reduce the message sizes that get communicated at each update round. Thirdly, in the case of FL, the server connects to all remote devices in the setting. Hence decentralized topologies act as an alternative solution during the instances of bottlenecks while operating with low bandwidth and high latency networks.

Federated Optimization: There exist various questions relevant to federated optimization in dealing with IID data. When employing intermittent communication graphs that capture the local SGD techniques, research gaps exist between the upper and lower bounds of optimization in FL scenarios. However, the convergence rates of these approaches do not tally with applicable lower constraints. Although the majority of the schemes can successfully achieve the asymptotically dominant statistical term, all of them fail to achieve the convergence rate of the accelerated mini-batch SGD. Hence there lies a scope of research that would enable federated averaging algorithms to eliminate the gap between the optimization bounds [153, 152, 154].

### 6.2. System and Data Heterogeneity

The capability to store, compute and communicate using the devices being part of a federated network vary significantly from one another. The reason behind such differences is related to hardware, network connection and power supply. Also, only a small part of the data should be active at any point in time. Here each device has the possibility of being unreliable because it is quite common for edge computing devices from the network due to connectivity or energy-based limitations. The non - identically distributed data collected from the various devices have a significant effect on the performance of FL systems. The issues pertinent to system heterogeneity can be achieved by implementing the following strategies. Firstly, the use of asynchronous communication to parallelize the functioning of iterative optimization algorithms. This also helps to eliminate struggles in a heterogeneous environment. Active sampling of devices at each round while aggregating device updates within the pre-defined window is a plausible solution to deal with heterogeneity. Ignoring fault tolerance and inclusion of algorithmic redundancy as part of code computation also helps to achieve algorithmic redundancy [153, 152, 154, 17].

**Possible Solution and Direction of Future Research** To avoid system heterogeneity, an asynchronous communication technique is used which helps to run the iterative optimization algorithms in parallel. This is a highly potential approach to eliminating the possibilities of stragglers in the heterogeneous environment. The second approach involves the selection of participating devices at each round actively to ensure maximum aggregation of updates within a pre-defined window wherein only a small subset of devices participate at the end of each round of training. The third plausible approach is to avoid device failures leading to bias in the device sampling scheme when the failed devices have some specific characteristics of data. Also, algorithmic redundancy could be introduced as part of a coded computation technique to achieve fault tolerance [153, 152, 154, 17].

### 6.3. Statistical Heterogeneity

As already mentioned, the various devices in the FL systems, collect data in a non-identically distributed manner across the network as per the varied types of uses by the end-users. The data point counts thus may vary significantly due to the presence of an underlying structure that captures the connectivity among the devices and their relevant distributions. The data generating paradigm increases the chances of having stragglers that add complexity to the modelling, analysis and evaluation of the system. The issues arise while training the federated models from the data that are non identically distributed both in terms of data modelling and analysis of the convergence behaviour relevant to associated training procedures [153, 152, 154, 17, 73, 74, 75].

**Possible Solution and Direction of Future Research**

Heterogeneous Diagnostics: There exist studies focusing on the quantification of statistical heterogeneity using different metrics which are mostly calculated during the training phase. This motivates the need for developing simplistic diagnostic techniques for the quantification of statistical and system heterogeneity even before the training phase. Also there lies scope of research to identify ways to improve the convergence techniques involved in federated optimization methodologies [153, 152, 154, 17, 73, 74, 75].

### 6.4. Privacy Issues

The privacy concerns persuade users to keep the raw data in each device localised in the federated settings. But sharing of various other information included in the training process has the potential to disclose sensitive information to the third party of the centralized server. The secure multiparty computation (SMC) or differential privacy techniques contributes toward achieving privacy in FL. But there are associated challenges of deteriorated performance and efficiency in the model. The management of balancing all of these security trade-offs acts as a significant challenge in achieving privacy in FL systems. In the case of FL, the model updates alone namely the gradient information is shared which ensures data protection on each device. But despite such attempts, there are open possibilities of sensitive information being shared with third parties or the centralized server while processing the updates of the communication model. Most of the development in this regard emphasizes the use of cryptography techniques or the implementation of differential privacy. But these techniques also have their challenges which motivate further development of innovative methodologies to achieve data privacy. As an example, some studies include a strategy in which all possible information about the counter party's data is acquired abiding by all the necessary protocols. There also exist techniques which allow the involvement of malicious parties where they can tamper with the system by feeding falsified input [153, 152, 154, 17].

**Possible Solutions and Future Direction of Research:**
The concept of privacy lies normally looked upon from the local or global level considering all devices in the network. The future direction of research lies in defining the concept of privacy at a much granular level as the various privacy-related constraints differ at various devices or even the data points in a single device. The objective lies in developing privacy techniques capable of handling mixed privacy restrictions [153, 152, 154, 17].

### 6.5. Security Issues

*Poisoning:*

There exist possibilities of two types of poisoning namely Data Poisoning and Model Poisoning in FL systems. Data poisoning in FL based training process occurs when multiple clients participate in the contribution of their on-device training data. In such circumstances, it becomes immensely difficult to stop or detect malicious clients from sending falsified fake data. These fake data poison the training process and the model also gets poisoned. In case of Model poisoning, the malicious client tampers the gradient or parameters which further modifies the received model before being sent to the centralized server for integration of information. The entire global model gets poisoned during this aggregation process where invalid gradients get included. With the increase in model dimensionality, the possibility of poisoning and backdoor attacks further increases. The plausible solution in this regard would be to share only prediction results or information which are less sensitive leading to the achievement of a more robust and protected method ensuring optimum level of privacy [153, 152, 154, 17, 73, 74, 75].

*Adversarial attacks:*

Adversarial attacks pose significant threats to the robustness and reliability of ML models, including those used in FL systems. These attacks involve manipulating input data to deceive models into making incorrect predictions. In [155] the authors provides valuable insights into strategies for mitigating such threats. By incorporating adversarial training and robust optimization techniques, the proposed framework aims to strengthen the defense mechanisms of models, ensuring their resilience in adversarial environments.

*Membership Inference Attacks:*

The raw data is usually stored in the local device yet there exists the chance of inferring the training data in an FL based system. As an instance, the information regarding the training data can be inferred during the process of updating the model in its learning process. The defence technique focuses on achieving differential privacy in this regard. The most common approach adopted is focused on ensuring secured computation, differential privacy and the development of a trusted execution environment. In the process of achieving secure computation, two main techniques are adopted namely - Secure Multiparty Computation (SMC) and Homomorphic Encryption. In the case of SMC, it is agreed to perform the inputs by two or more parties and the output is revealed to only a subset of the participants. On the contrary, in the case of homomorphic encryption, the computations are implemented on encrypted inputs before they get decrypted. In the process of implementing a differential privacy scheme, the user's contribution is masked. Noise is added to the clipped model parameters before integrating the model although it leads to compromising the model accuracy. The trusted execution environment presents a secured platform to run the FL process in low computation overhead which is presently suitable only for CPU devices [153, 152, 154, 17, 73, 74, 75].

Model Inversion Attacks: In this type of attack the training data gets compromised. The potential attacker gets access to the target labels and queries the final trained model. The returned classification scores get exploited to reconstruct the remaining data.

Membership Inference Attacks: Here the attackers try to find out if some data has been part of the training. Shadow models are created exploiting the returned classification scores with similar classification boundaries as the original model which was subjected to attack

Model Encoding Attacks: In this case, the attacker having white-box access attempts to find the training data which had been memorised by the model weights. On the contrary, in the case of a black-box situation, the original training model gets over-fitted for getting unauthorized access to parts of the training labels.

Model Stealing Attacks: In the case of a model stealing attack, the attacker intends to steal the entire model. When the model is sent to the participants for training, a duplicate model mimicking the original decision boundaries get created. Thus the attackers escape payment of usage fees to the ML experts of the original model or while selling the model to the third parties [153, 152, 154, 17, 73].

**Possible Solution and Direction of Future Research**

One of the possible mitigation steps to security concerns is the implementation of differential privacy schemes. In this case, the user's contribution before aggregating is masked by the addition of noise to the clipped model parameters. Although the accuracy of the model gets negatively affected due to the noise, it acts as a mathematical guarantee of achieving algorithmic output even when a particular data gets used for the training process [153, 152, 154, 17, 73].

The use of the Secure Multi-Party Computation (SMPC) function allows multiple participants to derive a process mutually similar to the ML training procedure. In such a case, only the outcome of the function gets disclosed to the participating members but not the relevant training information. The gradients and the parameters get computed, updated and encrypted through a decentralized process. The custody of each data item is split into shares that are sustained by the relevant participating entities. The SMPC helps in achieving privacy during the training period but remains vulnerable during the testing phase [153, 152, 154, 17, 73].

Homomorphic Encryption also acts as a mitigation technique wherein a complex cryptographic protocol is implemented that ensures mathematical derivation of the encrypted data and the generated outcome also remains encrypted. HE has the potential to ensure security in both the training and testing phase. The possible future direction of research lies in making this intensive computation technique economical to be fit for application in real-world scenarios [153, 152, 154, 17, 73].

### *6.6. Trust Issues*

In FL based frameworks, the ML model owners send their respective models to the data holders to implement training. This may seem to be secure but there exist trust issues. To highlight some of the examples, the ML model owners could alter their model and get information on their training data. Apart from this, in an FL setup, although the data is not shared in the original format, it has risks of being reconstructed in case there exists a lag in the protection of architecture and related parameters. There is also a possibility of FL exposing intermediate results namely the parameter updates from the optimization algorithms like stochastic gradient descent (SGD). The transmission of these gradients could also lead to the leaking of private information when it gets exposed to image pixel data structure. Hence protection of parameters remains to be an area of concern while striking a balance between privacy-security and achievement of system performance [153, 152, 154, 17, 73].

**Possible Solution and Direction of Future Research**

The potential solution to the aforementioned trust issues could be the implementation of the secure aggregator. Studies could be done in the development of an automated middleware procedure or program which would aggregate all the trained models and send the updates to the ML model owners acting as a robust model against various trust compromising attacks [153, 152, 154, 17, 73].

### *6.7. Algorithmic and Technical Challenges*

The traditional approaches do not work well in the FL paradigm which triggers the proposition of new algorithms such as Federated Averaging (FedAvg) and Federated Stochastic Variance Reduced Gradient (FSVRG) and the Co-Operative machine learning model. In FedAvg the entire training process is controlled by the centralised server which encompasses the shared global model enabling overall communication. Stochastic Gradient Decent Method is used for optimization to be done on the clients locally. In the case of the FSVRG technique, the full gradient computation is performed centrally and the other distribution updates are done at each client by performing random permutations on local data iteratively. The FedAvg and FSVRG approaches both focus on updating the model undergoing a synchronized approach for model updation. On the contrary, the Cooperative machine learning model takes an asynchronous approach which integrates the received client model with the global model. Although all the three techniques mentioned work well in the case of IID and non - IID but their potential in case of unevenly distributed data in a real-time environment is still unsure [153, 152, 154, 17, 73].

**Possible Solution and Direction of Future Research**

Development of Innovative Models of Asynchrony: The two predominant communication schemes explored in distributed optimization are the bulk synchronous and asynchronous techniques. On the contrary, in the case of federated networks, the devices often remain unassigned to the tasks in the process hence most devices remain inactive during any of the iterations. Hence as part of future research there lies the opportunity to develop device-centric communication models. These models work beyond the synchronous or asynchronous training schemes and individually each device decides on the time to interact with the server rather than being subjected to the workload [153, 152, 154, 17, 73].

Threshold level Communication Schemes: The knowledge of the optimum level or threshold level of communication required in FL is still unexplored. There lies opportunities for research to acquire detailed theoretical and empirical ideas relevant to one-shot and few-shot communication schemes in the implementations of massive statistically heterogeneous networks [153, 152, 154, 17, 73].

### *6.8. Autonomy*

Generally in an FL environment, all communication devices tend to be autonomous. In this regard, Association Autonomy deals with the autonomy and willingness of the devices to join or drop out of the network and also provides independence to participate in one or more networks. In the case of Communication Autonomy, the device has the freedom to decide on communicating with other participant devices and also the scale of data participation. Although FL is quite robust when communicating with any device while joining or leaving a system, the devices may also choose to make agreements to seamlessly enter or exit from an FL system. The Google FL system and blockchain are examples that contribute to the same purpose [153, 152, 154, 17, 73, 156].

**Possible Solution and Direction of Future Research** Adaptation of Centralized Training Workflows: The inclusion of centralized training workflows such as hyper-parameter tuning, neural architectures, debugging and interpretability in the FL setting acts as a bottleneck in the extensive adoption of FL in a real-time setting. Hence this acts as a potential area of future research. In future, FL-integrated machine and deep learning methodologies can process information at multi-granularity levels, working with multi-dimensional data, and building sustainable predictable models for better decision-making. [153, 152, 154, 17, 73].

## 7. Conclusions

Due to the unprecedented growth of IoT and generated data, there are many big data services and applications. However, conventional approaches using traditional AI/ML techniques for big data face critical issues, such as data privacy, data variety, communication efficiency, and scalability. In this context, FL is an AI breakthrough in the implementation of big data services and applications. As a result, this paper sets to provide a comprehensive survey on the use of FL for various big data services and applications. At the outset, fundamentals of big data and FL are discussed in association to the motivations for the integration of FL into big data. Then, literature pertaining to FL are reviewed which establishes FL as a promising AI technique to overcome the challenges in big data services and applications.This extensive review have enabled us to highlight the key challenges in FL implementation and guided towards potential research directions to overcome the same.

## Acknowledgement

The authors thank Praveen Kumar Reddy Maddikunta and Sweta Bhattacharya, SCORE, Vellore Institute of Technology for their valuable suggestions during the initial phases of the survey.

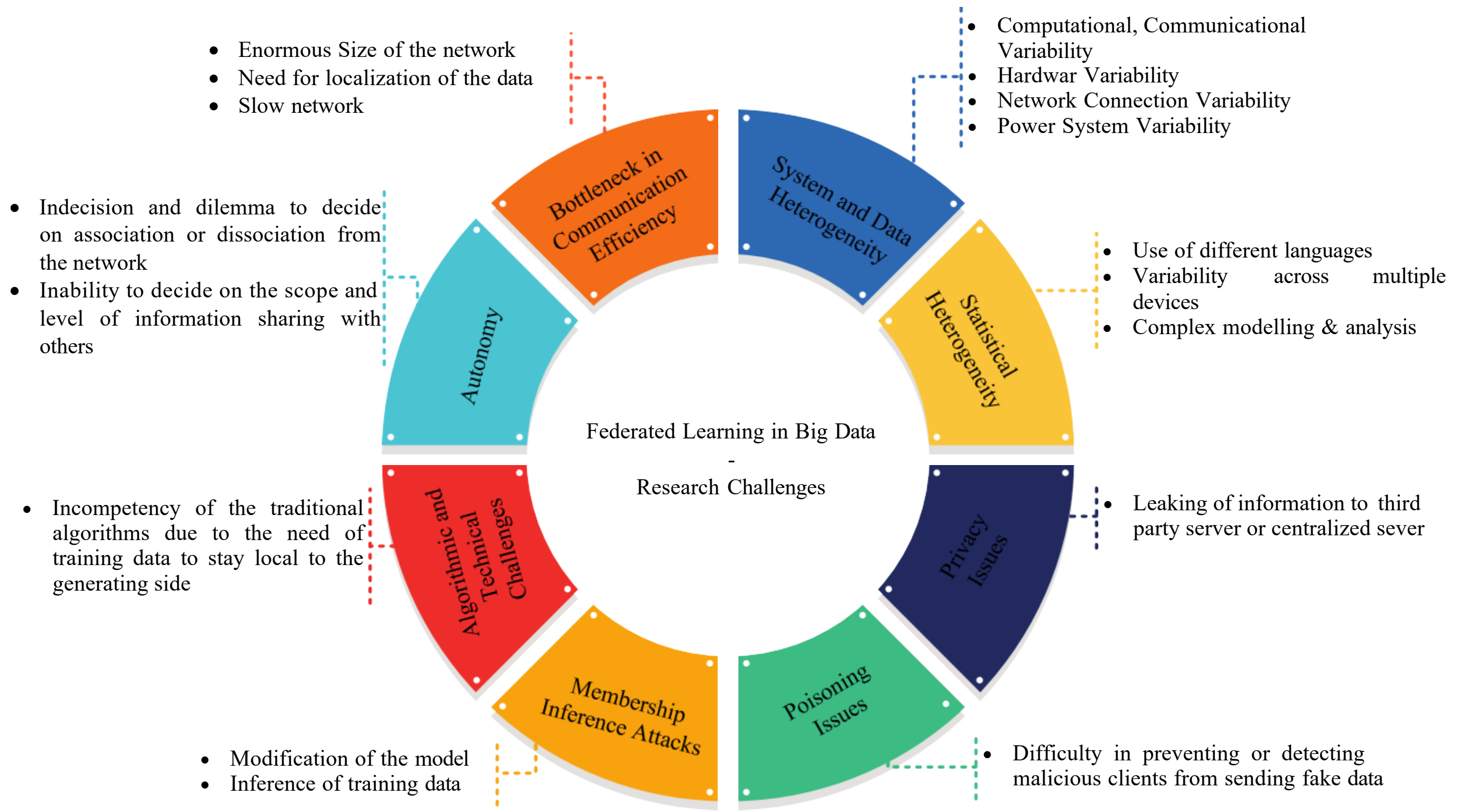

Figure 5: Federated learning in big data-research challenges.